%% file: emnlp2026-numpert-finetune.tex
\documentclass[11pt]{article}

\usepackage[final]{acl}

\usepackage{times}
\usepackage{latexsym}
\usepackage{booktabs}
\usepackage{pifont}
\usepackage{amsmath}
\usepackage{amssymb}
\usepackage{multirow}

\usepackage{arydshln}
\usepackage{nicematrix}

\newcommand{\xmark}{\ding{55}}
\definecolor{darkgreen}{RGB}{0,100,0}

\usepackage[T1]{fontenc}

\usepackage[utf8]{inputenc}

\usepackage{microtype}
\usepackage[T1]{fontenc}
\usepackage{amsmath}
\usepackage{tikz}
\usepackage{tcolorbox}
\usetikzlibrary{arrows.meta, positioning, calc}

\usepackage{inconsolata}
\usepackage{enumitem}

\usepackage{graphicx}

\newcommand{\qwenbase}{Qwen3-8B}

\newcommand{\original}{\emph{Original}}
\newcommand{\range}{\emph{Range}}
\newcommand{\numeration}{\emph{Numeration}}
\newcommand{\approximation}{\emph{Approximation}}
\newcommand{\random}{\emph{Random}}
\newcommand{\masked}{\emph{Masked}}
\newcommand{\negative}{\emph{Negative}}

\newcommand{\zeroshot}{Zero-Shot}
\newcommand{\twoshot}{Two-Shot}
\newcommand{\pap}{NCP}
\newcommand{\ncpfull}{Numerical Counterexample Prompt}

\newcommand{\ttt}{\texttt{T}$\rightarrow$\texttt{T}}
\newcommand{\ttf}{\texttt{T}$\rightarrow$\texttt{F}}
\newcommand{\ftf}{\texttt{F}$\rightarrow$\texttt{F}}

\newcommand{\groupone}{Base Group}
\newcommand{\grouptwo}{All Perturb Group}
\newcommand{\groupthree}{Single Perturb}
\newcommand{\groupfour}{Exclusion Group}

\newcommand{\grouponetable}{Base}
\newcommand{\grouptwotable}{All Perturb}
\newcommand{\groupthreetable}{Single Perturb}
\newcommand{\groupfourtable}{Exclusion}

\newcommand{\mgptfive}{$\text{M}_{\text{GPT-5.4}}$}
\newcommand{\mgemini}{$\text{M}_{\text{Gemini-2.5}}$}
\newcommand{\mbase}{$\text{M}_{\text{Base}}$}
\newcommand{\mbasetwoshot}{$\text{M}_{\text{Base-Two-Shot}}$}
\newcommand{\mbasepap}{$\text{M}_{\text{Base-NCP}}$}

\newcommand{\mallperturbsix}{$\text{M}_{\text{0.6B All-Perturbs}}$}
\newcommand{\mallperturb}{$\text{M}_{\text{All-Perturbs}}$}
\newcommand{\mallperturbseventeen}{$\text{M}_{\text{1.7B All-Perturbs}}$}
\newcommand{\mallperturbfour}{$\text{M}_{\text{4B All-Perturbs}}$}
\newcommand{\mbasethirtytwo}{$\text{M}_{\text{32B Base}}$}

\newcommand{\mptfulldata}{$\text{M}_{\text{PT-All-Perturbs}}$}

\newcommand{\mapproximation}{$\text{M}_{\text{Only-Approx}}$}
\newcommand{\mnumeration}{$\text{M}_{\text{Only-Num}}$}
\newcommand{\mrange}{$\text{M}_{\text{Only-Range}}$}
\newcommand{\mnegation}{$\text{M}_{\text{Only-Negation}}$}
\newcommand{\mrandom}{$\text{M}_{\text{Only-Random}}$}
\newcommand{\mmasked}{$\text{M}_{\text{Only-Masked}}$}
\newcommand{\mnoperturb}{$\text{M}_{\text{Only-Original}}$}
\newcommand{\mnooriginal}{$\text{M}_{\text{Excl-Original}}$}

\newcommand{\mnoapproximation}{$\text{M}_{\text{Excl-Approx}}$}
\newcommand{\mnonumeration}{$\text{M}_{\text{Excl-Num}}$}
\newcommand{\mnorange}{$\text{M}_{\text{Excl-Range}}$}
\newcommand{\mnonegation}{$\text{M}_{\text{Excl-Negation}}$}
\newcommand{\mnomasked}{$\text{M}_{\text{Excl-Masked}}$}
\newcommand{\mnorandom}{$\text{M}_{\text{Excl-Random}}$}

\newcommand{\mvitaminc}{$\text{M}_{\text{VitaminC}}$}
\newcommand{\mmixed}{$\text{M}_{\text{Mixed}}$}

\newcommand{\tcp}{TCP}
\newcommand{\tcpfull}{Threshold Comparison Prompt}
\newcommand{\scot}{SCoT}
\newcommand{\scotfull}{Structured Chain-of-Thought}

\title{Towards Robust Numerical Claim Verification} 

\author{
Peter Røysland Aarnes \\
University of Stavanger\\
\texttt{peter.r.aarnes@uis.no}
\And
Vinay Setty \\
Factiverse AI and University of Stavanger \\
\texttt{vsetty@acm.org}
}

\begin{document}
\maketitle
\begin{abstract}
Large language models (LLMs) are widely used for claim verification, yet remain brittle for numerical reasoning: even small changes in value can sharply degrade accuracy.
We show that this brittleness persists in frontier LLMs, but can be mitigated through adversarial fine-tuning on numerically perturbed examples.
Using parameter-efficient fine-tuning, small Qwen3 models (0.6B--8B) reach 98.7\% accuracy on label-flipping perturbations, outperforming larger zero-shot models and frontier systems (GPT-5.4 Pro (74.0\%) and Gemini 2.5 Flash (73.9\%)).
The gains generalise to unseen perturbation types, indicating robust numerical decision boundaries rather than memorised edits.
Robustness also transfers without target-domain data, significantly improving cross-lingual performance in Spanish.
We further show that the same fine-tuning recipe confers robustness to evidence-side perturbations, using the VitaminC dataset.
\end{abstract}

\input{emnlp2026-numpert-finetune-01}
\input{emnlp2026-numpert-finetune-02}
\input{emnlp2026-numpert-finetune-03}
\input{emnlp2026-numpert-finetune-04}
\input{emnlp2026-numpert-finetune-05-revised}
\input{emnlp2026-numpert-finetune-06}
\input{emnlp2026-numpert-finetune-07}
\newpage
\section*{Limitations}

\paragraph{Data and perturbation scope.} Our perturbation taxonomy rewrites numerical expressions in the claim while leaving evidence unchanged. Section~\ref{sec:vitaminc} reaches the evidence side through VitaminC's human-authored revisions, but our generator does not produce them, and jointly perturbing claim and evidence remains an open direction. The six types focus on scalar quantities as digits or written-out numbers, so robustness to unit conversions, date arithmetic, compound quantities, or numbers in tables is not guaranteed. Training and primary evaluation use QuanTemp (English fact-checking); cross-dataset experiments add Spanish, AVeriTeC and VitaminC but remain in English and Spanish and, apart from VitaminC's single-sentence evidence, within the fact-checking domain, so broader language coverage and generalisation to scientific, financial, or tabular settings both require further evaluation.

\paragraph{Model and evaluation setup.} Experiments use only the Qwen3 family (0.6B--32B) with QLoRA and soft prompt tuning. \citet{Aarnes:2025:IJCNLP} document the same numerical brittleness on LLaMA-family models, so we expect the qualitative findings to hold across LLaMA, Mistral, Gemma, and comparable open families, though confirming this directly is left for future work; quantitative effects (the scale at which diminishing returns set in, the size of the QLoRA-vs.-soft-prompt gap) may shift across families and PEFT methods. The soft-prompt baseline reuses QLoRA hyperparameters, making it a matched-budget rather than optimally tuned comparison. All experiments assume gold evidence, isolating verification from retrieval; whether the robustness gains survive noisy or incomplete retrieval is left for future work.

\paragraph{Interpretability.} We lack a mechanistic account of \emph{why} adversarial fine-tuning works. It is unclear whether models learn to attend more carefully to numerical tokens, develop surface-form-invariant representations, or rely on other signals; attribution, probing, and activation-patching analyses would be needed to explain the gains and anticipate failure modes --- a direction we intend to pursue.

\section*{Ethical Considerations}

This work studies the robustness of LLMs used for numerical claim verification. We identify specific weaknesses that existing models exhibit when claims contain perturbed numerical expressions, and we show that adversarial fine-tuning substantially reduces these vulnerabilities. We do not endorse any specific model or configuration for deployment in real-world fact-checking pipelines; fact-checking is a nuanced task where model outputs should be treated as decision-support tools rather than authoritative verdicts, and human oversight remains essential.

The perturbation techniques studied could, in principle, be used to craft adversarial claims that evade automated fact-checkers. We consider this a reason to publish rather than withhold the findings: understanding and documenting these failure modes is a prerequisite for building more resilient systems, and the perturbation types we study are straightforward enough that they are already available to any motivated adversary. All data used in this work is sourced from publicly available fact-checking corpora, and no personal or sensitive information is involved.

\section*{Acknowledgements}
This research is funded by SFI MediaFutures partners and the Research Council of Norway (grant number 309339).

\bibliography{emnlp2026-numpert-finetune}

\input{emnlp2026-numpert-finetune-08-APPENDIX-A-revised}

\end{document}

%% file: emnlp2026-numpert-finetune-01.tex
\section{Introduction}
\label{sec:intro}

Claim verification is an important societal task, and LLMs are a popular choice. Claims that contain numerical values such as statistics, counts, percentages, and dates are especially challenging to verify, even for humans~\cite{Aly:2021:NURIPS} and for frontier large language models~\cite{V:2024:Sigir,Aarnes:2025:IJCNLP, Akhtar:2023:ACL}, as they require careful interpretation of context (time spans, units, denominators) and cross-checking against reliable sources. Figure~\ref{fig:perturbation_example} illustrates the problem: a simple surface rewrite that changes ``3.5\%'' to ``six percent'' flips the correct label, yet a zero-shot LLM misses it.

\begin{figure}[t!]
\begin{tcolorbox}[title=Numerical Perturbation Example,
left=1pt, colback=white]
\footnotesize
\begin{tabular}{p{\columnwidth}}

\textbf{Original Claim:} \emph{``The US unemployment rate fell to \textcolor{darkgreen}{3.5\%} in February 2020, matching the lowest rate in five decades.''} \\[0.3em]

[\textbf{Label:} \texttt{TRUE},\quad \textbf{Zero-shot: \textcolor{darkgreen}{\texttt{TRUE}}}~$\checkmark$,\quad \textbf{Fine-tuned: \textcolor{darkgreen}{\texttt{TRUE}}}~$\checkmark$] \\[0.5em]

\textbf{Perturbed Claim:} \emph{``The US unemployment rate fell to \textcolor{red}{six percent} in February 2020, matching the lowest rate in five decades.''} \\[0.3em]

[\textbf{Label:} \texttt{FALSE},\quad \textbf{Zero-shot: \textcolor{red}{\texttt{TRUE}}}~\xmark,\quad \textbf{Adversarially Fine-tuned: \textcolor{darkgreen}{\texttt{FALSE}}}~$\checkmark$] \\[0.5em]

\textbf{Evidence:} \emph{``(...) According to the U.S. Bureau of Labor Statistics: `The unemployment rate was \textcolor{darkgreen}{3.5 percent} in February 2020, unchanged from January 2020 (...)'\,''}
\vspace{-10pt}
\end{tabular}
\end{tcolorbox}
\caption{\footnotesize Numeration perturbation: rewriting ``3.5\%'' as ``six percent'' flips the label to \texttt{FALSE}. Zero-shot predicts \texttt{TRUE} (wrong); adversarial fine-tuning predicts \texttt{FALSE} (correct).}
\label{fig:perturbation_example}
\end{figure}

Despite strong performance on arithmetic or structured quantitative tasks~\cite{Fang:2025:CS,Schwartz:2024:EMNLP}, LLM robustness remains uneven in realistic verification settings. Recent works suggest that even frontier models can break down when the numerical values in a claim are slightly perturbed, revealing brittleness and sensitivity that can undermine trust in their outputs~\cite{Akhtar:2023:ACL, Aarnes:2025:IJCNLP, Yang:2025:EACL}. This brittleness is particularly problematic in evidence-grounded verification, where a model must reconcile a claim's surface form (e.g., a written-out number) with the evidence's surface form (e.g., a digit), and anchoring on lexical overlap rather than semantic equivalence is an easy failure mode. This gap highlights that strong performance on benchmark-style numerical reasoning does not automatically translate into robust, real-world verification of numerical claims.

In this paper, we investigate whether adversarial fine-tuning (training on numerically perturbed claims) can improve robustness in numerical claim verification. We apply six perturbation categories (numeration, approximation, range, negation, masking, and random replacement) to the QuanTemp dataset~\cite{V:2024:Sigir}, fine-tune Qwen3 models from 0.6B to 8B parameters using QLoRA and soft prompt tuning, and evaluate cross-dataset generalisation to English and Spanish corpora unseen during training, addressing four research questions:

\begin{description}[leftmargin=0pt,labelindent=0pt]
    \item[\textbf{RQ1}]: \textit{Does adversarial fine-tuning improve robustness to label-flipping perturbations over clean-data fine-tuning?}
    \item[\textbf{RQ2}]: \textit{Does training on a perturbation subset generalise, and how does training composition affect robustness?}
    \item[\textbf{RQ3}]: \textit{How do model scale and adaptation method (QLoRA vs.\ soft prompts) interact with adversarial fine-tuning?}
    \item[\textbf{RQ4}]: \textit{Does adversarial fine-tuning generalise across domains and languages?}
\end{description}

Our experiments show that adversarial fine-tuning is highly effective: a model trained on clean data alone collapses to under 13\% accuracy on label-flipping perturbations, while QLoRA fine-tuning on all six perturbation types restores accuracy to 96--100\%. Training composition matters more than variety; fine-tuned small models (0.6B) surpass zero-shot 32B models on the hardest perturbations. Crucially, adversarially fine-tuned Qwen3-8B models also substantially outperform frontier closed-source models (GPT-5.4 Pro, Gemini 2.5 Flash) -- for instance, achieving 97.7\% vs.\ 74.0\% micro-average accuracy on label-flipping perturbations. Cross-dataset generalisation is practically robust: fine-tuned models outperform the zero-shot baseline on Spanish (97.5\% vs.\ 96.4\%), whereas on AVeriTeC the accuracy difference is not significant and the gain is in coverage (the baseline returns no parseable verdict on 37.5\% of instances). Detailed answers to each RQ are in Section~\ref{sec:results}.

Our main contributions are: \textbf{(i)} a systematic study of LLM brittleness to numerical claim perturbations and its recovery by fine-tuning on adversarial (counterfactually constructed) examples; \textbf{(ii)} ablations showing which perturbation types are \emph{essential} (\masked, \numeration) and which are redundant (\approximation, \random); \textbf{(iii)} the counterintuitive finding that unperturbed training examples, rather than the perturbations themselves, shift the decision boundary and reduce label-preserving accuracy; \textbf{(iv)} evidence that adversarially fine-tuned small models (0.6B--8B) substantially outperform zero-shot 32B and frontier closed-source models (GPT-5.4 Pro, Gemini 2.5 Flash) on label-flipping perturbations; and \textbf{(v)} the first cross-lingual evaluation showing numerical robustness transfers from English to Spanish without target-language training data.

%% file: emnlp2026-numpert-finetune-02.tex
\section{Related work}
\label{sec:related}

Benchmarks such as FEVER~\citep{Thorne:2018:NAACL} and methods from retrieval pipelines to program-guided reasoning~\citep{Guo:2022:TACL,Pan:2023:ACL} are popular. Numerical claims are a particularly hard sub-problem~\citep{V:2024:Sigir,Lu:2023:EMNLP}; we build on QuanTemp~\cite{V:2024:Sigir}, and the perturbation taxonomy of \citet{Aarnes:2025:IJCNLP}.

LLM robustness to numerical perturbations is fragile. While targeted training and better encodings reduce numerical errors~\citep{Wallace:2019:EMNLP,Geva:2020:ACL,Schwartz:2024:EMNLP}, small value changes still flip predictions across QA~\citep{Akhtar:2023:ACL,Xu:2022:EMNLP,Zhou:2024:ACL}, claim verification~\citep{Aarnes:2025:IJCNLP,Mamta:2025:NAACL}, and mathematical reasoning~\citep{Yang:2025:EACL}. The brittleness is compounded in multilingual settings, where surface-form variation across languages (e.g., decimal separators, numeral systems, written-out forms) creates additional attack surface that monolingual robustness studies have not addressed~\citep{Alam:2025:CLEF}.

Adversarial training addresses this brittleness: worst-case perturbation augmentation~\citep{Madry:2018:ICLR}, gradient-based NLP perturbations~\citep{Guo:2021:EMNLP}, and math-problem adversaries~\citep{Zhou:2024:AAAI} all improve robustness, as do counterfactually revised~\citep{Kaushik:2020:ICLR} and adversarially collected~\citep{Nie:2020:ACL} training sets. We apply this strategy to numerical claim verification across six perturbation types.

Parameter-efficient fine-tuning makes large-model adaptation practical via LoRA~\citep{Hu:2022:ICLR}, QLoRA~\citep{Dettmers:2023:NIPS}, and soft prompt tuning~\citep{Lester:2021:EMNLP}. Fine-tuned models consistently outperform larger zero-shot LLMs on fact-checking~\citep{Setty:2024:SIGIR,Tang:2024:EMNLP,Larsson:2026:ECIR}. To the best of our knowledge, there are no prior studies applying adversarial fine-tuning for numerical perturbations in claim verification setting.

%% file: emnlp2026-numpert-finetune-03.tex
\section{Methodology}
\label{sec:Methodology}

Our method consists of two main steps: First, we generate adversarial perturbations for numerical claims, where we adopt a similar taxonomy and methodology to \citet{Akhtar:2023:ACL, Aarnes:2025:IJCNLP}. Second, we conduct adversarial fine-tuning on a set of models, using a mix of perturbed and original data.

\subsection{Generating Numerical Perturbations}
\label{sec:dataset}
We use the QuanTemp dataset~\cite{V:2024:Sigir} of real-world numerical claim--evidence pairs, with evidence verdict summaries removed and labels binarised to \texttt{True}/\texttt{False}.

\paragraph{Dataset and Perturbation Types.}

\input{tables/dataset-stats/dataset-stats-test.tex}

Numerical values in claims are automatically tagged by a spaCy\footnote{\url{https://spacy.io/api/entityrecognizer}} named entity tagger model. We apply the six perturbation types defined by \citet{Aarnes:2025:IJCNLP} to the numerical values (see Appendix~\ref{app:perturbation_types} for definitions and examples): \textbf{Numeration} (digit$\leftrightarrow$word form, e.g.\ ``350''$\to$``three hundred fifty''); \textbf{Approximation} (insert ``about''); \textbf{Range} (exact value $\to$ numerical range); \textbf{Random} (substitute a random value of equal character length); \textbf{Negation} (invert the value); \textbf{Masking} (replace the value with ``\#'' symbols of equal character length). For numeration, approximation, and range, we generate both a value-preserving and a value-changing variant, which are label-preserving and label-flipping for \texttt{True} claims, while both remain \texttt{False} for \texttt{False} claims; random, negation, and masking produce only \texttt{False} variants. Of note, masking differs from the other types by making a claim false through the omission of its numerical values. To prevent cross-split leakage, perturbations are generated within each predefined QuanTemp split, so perturbed variants of a claim remain in the split of their source. Table~\ref{tab:dataset_overview_test} reports the statistics for the test split; statistics for the training and validation splits are provided in Appendix~\ref{app:dataset_stats}.

\paragraph{Cross-Dataset Generalisation Splits.}
\label{sec:generalization_datasets}

We evaluate generalisation on two held-out datasets: \textbf{AVeriTeC 2.0} dev~\citep{Schlichtkrull:2023:NeurIPs,Akhtar:2025:FEVER} (119 numerical claims, 631 perturbed instances) and a \textbf{Spanish} test set from CLEF CheckThat! 2025~\cite{Venktesh:2025:CLEF} (2{,}633 claims, 13{,}827 perturbed instances). Dataset statistics and construction details are in Appendix~\ref{app:crossdata}.

\subsection{Adversarial Fine-tuning}

\input{tikz_diagram}

We frame numerical claim verification as binary classification: given a claim $x_c$ and evidence $x_e$, a model predicts $\hat{y} \in \{\texttt{True}, \texttt{False}\}$. Adversarial fine-tuning augments the standard training set $\mathcal{D}_{\text{orig}} = \{(x_c^{(i)}, x_e^{(i)}, y^{(i)})\}$ with perturbed instances $\mathcal{D}_{\text{adv}} = \{(\tilde{x}_c^{(i,t)}, x_e^{(i)}, \tilde{y}^{(i,t)})\}$, where $\tilde{x}_c^{(i,t)}$ is claim $i$ perturbed under type $t$ and $\tilde{y}^{(i,t)}$ is the revised label. The full training mixture $\mathcal{T} = \mathcal{D}_{\text{orig}} \cup \mathcal{D}_{\text{adv}}$ is optimised under binary cross-entropy:
\begin{equation*}
  \mathcal{L}(\mathcal{T}) = -\frac{1}{|\mathcal{T}|} \sum_{(x,\,y)\,\in\,\mathcal{T}} \bigl[ y \log \hat{y} + (1 - y) \log(1 - \hat{y}) \bigr]
\end{equation*}

We compare two independent parameter-efficient adaptation strategies (Figure~\ref{fig:arch}), both keeping the pre-trained backbone $\Theta$ frozen.

\textbf{QLoRA}~\citep{Hu:2022:ICLR,Dettmers:2023:NIPS}. The backbone $\Theta$ is loaded in 4-bit NormalFloat (NF4) precision. A low-rank residual is inserted into each selected projection matrix $W^{(l)} \in \mathbb{R}^{d \times k} : 
  W^{(l)} \leftarrow W^{(l)} + B^{(l)} A^{(l)}, 
  A^{(l)} \in \mathbb{R}^{r \times k}, B^{(l)} \in \mathbb{R}^{d \times r}
$
with rank $r \ll \min(d, k)$. $A^{(l)}$ is initialised from a Gaussian distribution and $B^{(l)}$ is initialised to zero, so $\Delta W^{(l)} = 0$ at the start of training. Only $\{A^{(l)}, B^{(l)}\}$ and the veracity head are trained. Adapters are inserted into all attention ($Q$, $K$, $V$, $O$) and feed-forward (gate, up, down) projection matrices with rank $r = 32$, implemented via the Unsloth\footnote{\url{https://unsloth.ai/}} framework.

\textbf{Soft prompt tuning}~\citep{Lester:2021:EMNLP} prepends $m$ (set to 20) learnable continuous vectors $P \in \mathbb{R}^{m \times d}$ to the frozen input embeddings, training only $P$ and a veracity head while all backbone weights remain fixed. For the full formulation, see Appendix~\ref{app:soft_prompt}.

\subsubsection{Training Data Sampling}

We use four training configurations: (1) original only; (2) leave-one-out (all perturbation types except one); (3) single perturbation type; and (4) all perturbation types. Each is balanced 1:1 between original and perturbed examples; the validation set retains its natural distribution.

\subsection{Evaluation and Fine-tuning}
A mid-sized base model serves both as the primary model for fine-tuning and as the baseline for different prompting strategies. We also evaluate a larger base model variant and two proprietary models in a zero-shot setting to establish an upper-bound reference point. Lastly, three smaller model variants are fine-tuned with all perturbations and original data for traditional adversarial training (see Section~\ref{sec:experimental_setup} for the specific models used). Full code and data can be accessed through our GitHub repository\footnote{\url{https://github.com/iai-group/numpert-fine-tune}}.

\subsection{Prompting Strategy}
All models use the same system and instruction prompts for fine-tuning and evaluation. Besides the \textit{Zero-shot} setup, the base model is also evaluated with a \textit{Two-shot} strategy, including one \texttt{True} and one \texttt{False} claim--evidence pair. These prompts are provided in Appendix~\ref{app:prompts}. We also tested \ncpfull\ (NCP), but omit it from the main evaluation due to high parse-failure rates (42--61\%; details are found in Appendix~\ref{app:ncp}).

%% file: tables/dataset-stats/dataset-stats-test.tex

\begin{table}[t!]
\small
\centering
\resizebox{\columnwidth}{!}{%
\begin{tabular}{lcccc}

\toprule
\textbf{Category} & \textbf{T$\rightarrow$T} & \textbf{T$\rightarrow$F} & \textbf{F$\rightarrow$F} & \textbf{Total} \\
\midrule
\textbf{\numeration} & 216 & 216 & 1002 & 1434 \\
\textbf{\approximation} & 139 & 139 & 784 & 1062 \\
\textbf{\range} & 190 & 190 & 843 & 1223 \\
\textbf{\masked} & \xmark & 216 & 501 & 717 \\
\textbf{\random} & \xmark & 216 & 501 & 717 \\
\textbf{\negative} & \xmark & 52 & 91 & 143 \\
\bottomrule
\end{tabular}%
}
\vspace{-5pt}
\caption{\footnotesize QuanTemp \textbf{test} split: sample counts by perturbation type and transition direction. Original claims: 264 \texttt{True}, 615 \texttt{False}.}
\label{tab:dataset_overview_test}
\end{table}

%% file: tikz_diagram.tex

\begin{figure}[t!!]
    \centering
    \resizebox{0.4\textwidth}{!}{%
    \begin{tikzpicture}[
        font=\sffamily,
        box/.style={draw, thick, rounded corners, align=center, minimum width=3.3cm, minimum height=1cm},
        bigbox/.style={draw, thick, rounded corners, align=center, minimum width=11cm, fill=gray!5, inner ysep=8pt},
        arrow/.style={thick, -{Stealth[length=2mm]}}
    ]
    \node[box, fill=blue!10, text width=4cm, align=center] (input_pair)
    {\footnotesize Numerical Claim and Evidence \\[-1pt]\scriptsize $(x_c, x_e)$ or perturbed $(\tilde{x}_c, x_e)$};

    \node[box, fill=gray!20, text width=2.5cm, minimum width=3cm, above=1.1cm of input_pair, xshift=-1.8cm, align=center] (embed)
    {\footnotesize Input Embedding \\[-1pt]\scriptsize $E(x_c, x_e)$ (frozen)};

    \node[box, fill=green!25, text width=2.5cm, minimum width=3cm, above=1.1cm of input_pair, xshift=1.8cm, align=center] (prompt)
    {\footnotesize \textbf{(a)} Soft Prompt \\[-1pt]\scriptsize Trainable $P \in \mathbb{R}^{m \times d}$};

    \coordinate (split) at ([yshift=0.45cm]input_pair.north);
    \draw[thick] (input_pair.north) -- (split);
    \draw[thick] (split -| embed.south) -- (split -| prompt.south);
    \draw[arrow] (split -| embed.south) -- (embed.south);
    \draw[arrow] (split -| prompt.south) -- (prompt.south);

    \node[bigbox, minimum width=7.5cm, above=2.8cm of input_pair] (main) {
        \begin{minipage}{7cm}
            \centering
            \textbf{\footnotesize Frozen transformer $\Theta$ with \textbf{(b)} QLoRA adapters $\{A^{(l)}, B^{(l)}\}$}\\[4pt]
            \begin{tikzpicture}[every node/.style={box}]
                \node[fill=orange!35, text width=2.5cm, minimum width=2.5cm, align=center] (tb)
                {\footnotesize Transformer Block\\[-1pt]\scriptsize Frozen, 4-bit $\Theta$};
                \node[fill=green!25, text width=2.5cm, minimum width=2.5cm, align=center, right=0.5cm of tb]
                {\footnotesize \textbf{(b)} QLoRA Adapter\\[-1pt]\scriptsize Trainable $\{A^{(l)}, B^{(l)}\}$};
            \end{tikzpicture}
        \end{minipage}
    };

    \draw[arrow] (embed.north) -- (main.south -| embed.north);
    \draw[arrow] (prompt.north) -- (main.south -| prompt.north);

    \node[align=center] at ($(embed.north)!0.5!(prompt.north) + (0,0.55)$)
    {\scriptsize $h^{(0)} = [\,P\,;\,E(x_c, x_e)\,]$};

    \node[box, fill=gray!10, above=0.6cm of main, text width=5cm] (head)
    {\footnotesize Veracity Head\\[-1pt]\scriptsize $\hat{y} = \sigma(\mathbf{w}^{\top} h^{(L)}_{\text{cls}} + b)$,\; $\hat{y} \in [0,1]$};
    \draw[arrow] (main.north) -- node[right, xshift=2pt]{\scriptsize $h^{(L)}$} (head.south);

    \node[box, fill=pink!20, text width=6.5cm, above=0.7cm of head] (loss)
    {\footnotesize Training Objective \\[-1pt]\scriptsize $\mathcal{L}(\mathcal{T}) = -\tfrac{1}{|\mathcal{T}|} \sum\limits_{(x,y) \in \mathcal{T}} \big[ y \log \hat{y} + (1-y) \log(1-\hat{y}) \big]$};
    \draw[arrow] (head.north) -- (loss.south);

    \end{tikzpicture}%
    }
    \caption{\footnotesize Two PEFT strategies for numerical claim verification. \textbf{(a) Soft prompt tuning} prepends $m$ learnable vectors $P$ to frozen embeddings; \textbf{(b) QLoRA} inserts low-rank residuals into a 4-bit quantized backbone $\Theta$. Both share a sigmoid veracity head trained on mixture $\mathcal{T}$ of original and perturbed examples.}
    \label{fig:arch}
\end{figure}
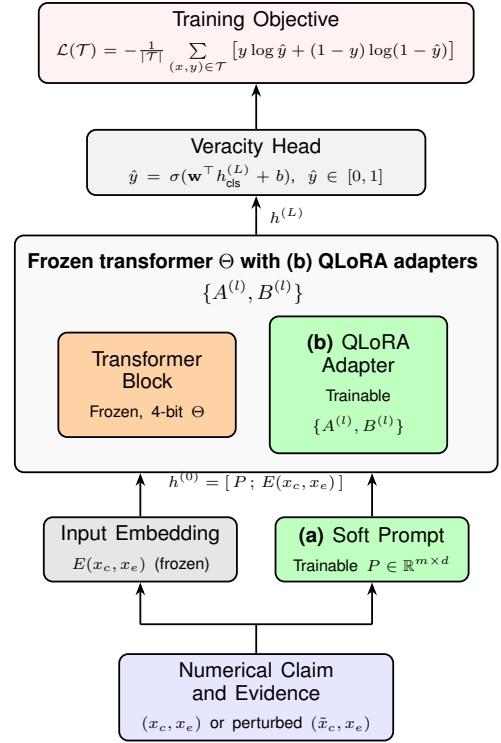

%% file: emnlp2026-numpert-finetune-04.tex
\section{Experimental Setup}
\label{sec:experimental_setup}

\subsection{Model Set and Inference Settings}

We use Qwen3-8B as the primary model, with additional experiments across Qwen3 0.6B, 1.7B, 4B, and 32B (Unsloth Q4\_K\_M GGUF quantisation~\citep{Dettmers:2023:NIPS}). Models run at $0.001$ temperature, with thinking enabled, and use the prompting strategy described in Section~\ref{sec:Methodology}. The Qwen3 family is selected because it provides openly available checkpoints from 0.6B to 32B under a consistent pre-training recipe, enabling controlled scale ablations within a single family. \citet{Aarnes:2025:IJCNLP} report the same numerical brittleness on LLaMA-family models for the underlying claim verification task, so we expect our qualitative findings to reflect the perturbation taxonomy and verification task rather than Qwen3-specific properties; quantitative effect sizes may differ across families. Results for unquantised models can be found in Appendix~\ref{app:unquantized_results}.

As closed-source frontier baselines we evaluate \mgptfive\ (GPT-5.4 Pro via the Azure OpenAI API, version \texttt{2024-12-01-preview}) and \mgemini\ (\texttt{gemini-2.5-flash}, knowledge cutoff January 2025), both queried zero-shot with the same prompt template. 

\subsection{Parameter-Efficient Adaptation}
\label{sec:peft}

Models are trained for 2 epochs with a 1:1 ratio of original to perturbed examples, using AdamW (8-bit) with learning rate $1\times10^{-4}$, cosine schedule, and a maximum sequence length of 16{,}384 tokens. Full hyperparameters are listed in Appendix~\ref{app:hyperparams}. For soft prompt tuning we set $m=20$, while keeping all other hyperparameters the same as in the QLoRA setting.
\input{tables/results_q4_k_m/quantemp_results/T-_F}
\subsection{Evaluation}

We report accuracy per perturbation type (relative to the corresponding original examples) under each evaluation setting on the test split of the QuanTemp dataset explained in Section~\ref{sec:dataset}. We use strict scoring: only outputs that match the required JSON schema (\texttt{\{"label": true/false\}}) are included; all others are counted as parse failures and excluded (Appendix~\ref{app:parsing_errors}). Parse failures are largely confined to the non-fine-tuned baseline and depend on prompting, whereas fine-tuned models exhibit near-zero failures. Thus, counting parse failures as incorrect would mainly lower baseline accuracy, with negligible effect on fine-tuned results.

All experiments use gold evidence to isolate the contribution of adversarial fine-tuning from retrieval quality. This controlled setup attributes observed performance differences unambiguously to the verification component rather than confounding them with retrieval errors; combining these gains with imperfect retrieval is an orthogonal question.

We report single-run results, as repeating inference across our $>$20 model configurations would be prohibitively expensive. The QuanTemp effects we discuss are large and consistent across perturbation types and evaluation directions, supporting confidence in the qualitative conclusions; for the cross-dataset, distance, and VitaminC comparisons, we test significance on the stored per-instance predictions (Sections~\ref{sec:significance}--\ref{sec:vitaminc}).

%% file: tables/results_q4_k_m/quantemp_results/T-_F.tex
\begin{table*}[h!!]
\centering
\definecolor{numpertdeltaimprove}{RGB}{0,100,0}
\definecolor{numpertdeltadecline}{RGB}{180,35,24}
\small
\setlength{\tabcolsep}{3pt}
\resizebox{\textwidth}{!}{%
\begin{NiceTabular}{l|c:ccccccc}[first-col]
\toprule
 & \textbf{Model} & \textbf{Orig.} & \textbf{Approx.} & \textbf{Num.} & \textbf{Range} & \textbf{Neg.} & \textbf{Mask.} & \textbf{Rand.} & \textbf{M. Avg.} \\
\midrule
\Block{6-1}{\rotate \grouponetable}
& \mgptfive & 78.12 & 92.70\textsuperscript{\textcolor{numpertdeltaimprove}{+14.58}} & 86.06\textsuperscript{\textcolor{numpertdeltaimprove}{+7.94}} & 93.44\textsuperscript{\textcolor{numpertdeltaimprove}{+15.32}} & 39.22\textsuperscript{\textcolor{numpertdeltadecline}{-38.90}} & 17.00\textsuperscript{\textcolor{numpertdeltadecline}{-61.12}} & 90.14\textsuperscript{\textcolor{numpertdeltaimprove}{+12.02}} & 73.96\textsuperscript{\textcolor{numpertdeltadecline}{-4.16}} \\
& \mgemini & 74.71 & 91.97\textsuperscript{\textcolor{numpertdeltaimprove}{+17.26}} & 86.05\textsuperscript{\textcolor{numpertdeltaimprove}{+11.34}} & 92.43\textsuperscript{\textcolor{numpertdeltaimprove}{+17.72}} & 53.85\textsuperscript{\textcolor{numpertdeltadecline}{-20.86}} & 19.62\textsuperscript{\textcolor{numpertdeltadecline}{-55.09}} & 92.06\textsuperscript{\textcolor{numpertdeltaimprove}{+17.35}} & 73.91\textsuperscript{\textcolor{numpertdeltadecline}{-0.80}} \\
& \mbase & 86.35 & 85.05\textsuperscript{\textcolor{numpertdeltadecline}{-1.30}} & 69.59\textsuperscript{\textcolor{numpertdeltadecline}{-16.76}} & 82.35\textsuperscript{\textcolor{numpertdeltadecline}{-4.00}} & 27.27\textsuperscript{\textcolor{numpertdeltadecline}{-59.08}} & 8.87\textsuperscript{\textcolor{numpertdeltadecline}{-77.48}} & 86.98\textsuperscript{\textcolor{numpertdeltaimprove}{+0.63}} & 62.80\textsuperscript{\textcolor{numpertdeltadecline}{-23.55}} \\
 & \mbasetwoshot & 81.96 & 88.72\textsuperscript{\textcolor{numpertdeltaimprove}{+6.76}} & 73.21\textsuperscript{\textcolor{numpertdeltadecline}{-8.75}} & 84.78\textsuperscript{\textcolor{numpertdeltaimprove}{+2.82}} & 28.85\textsuperscript{\textcolor{numpertdeltadecline}{-53.11}} & 12.44\textsuperscript{\textcolor{numpertdeltadecline}{-69.52}} & 90.61\textsuperscript{\textcolor{numpertdeltaimprove}{+8.65}} & 66.10\textsuperscript{\textcolor{numpertdeltadecline}{-15.86}} \\
 & \mbasethirtytwo & 85.55 & 88.64\textsuperscript{\textcolor{numpertdeltaimprove}{+3.09}} & 78.89\textsuperscript{\textcolor{numpertdeltadecline}{-6.66}} & 86.67\textsuperscript{\textcolor{numpertdeltaimprove}{+1.12}} & 36.00\textsuperscript{\textcolor{numpertdeltadecline}{-49.55}} & 16.19\textsuperscript{\textcolor{numpertdeltadecline}{-69.36}} & 86.41\textsuperscript{\textcolor{numpertdeltaimprove}{+0.86}} & 67.89\textsuperscript{\textcolor{numpertdeltadecline}{-17.66}} \\
 & \mnoperturb & 95.83 & 8.63\textsuperscript{\textcolor{numpertdeltadecline}{-87.20}} & 7.41\textsuperscript{\textcolor{numpertdeltadecline}{-88.42}} & 9.47\textsuperscript{\textcolor{numpertdeltadecline}{-86.36}} & 7.69\textsuperscript{\textcolor{numpertdeltadecline}{-88.14}} & 5.56\textsuperscript{\textcolor{numpertdeltadecline}{-90.27}} & 12.50\textsuperscript{\textcolor{numpertdeltadecline}{-83.33}} & 8.65\textsuperscript{\textcolor{numpertdeltadecline}{-87.18}} \\
\midrule
\Block{4-1}{\rotate \grouptwotable} & \mptfulldata & 81.71 & 40.00\textsuperscript{\textcolor{numpertdeltadecline}{-41.71}} & 32.06\textsuperscript{\textcolor{numpertdeltadecline}{-49.65}} & 39.78\textsuperscript{\textcolor{numpertdeltadecline}{-41.93}} & 26.53\textsuperscript{\textcolor{numpertdeltadecline}{-55.18}} & 18.32\textsuperscript{\textcolor{numpertdeltadecline}{-63.39}} & 52.38\textsuperscript{\textcolor{numpertdeltadecline}{-29.33}} & 35.80\textsuperscript{\textcolor{numpertdeltadecline}{-45.91}}\\
 & \mallperturbsix & 84.85 & 87.05\textsuperscript{\textcolor{numpertdeltaimprove}{+2.20}} & 76.85\textsuperscript{\textcolor{numpertdeltadecline}{-8.00}} & 78.95\textsuperscript{\textcolor{numpertdeltadecline}{-5.90}} & \textbf{100.00}\textsuperscript{\textcolor{numpertdeltaimprove}{+15.15}} & 99.54\textsuperscript{\textcolor{numpertdeltaimprove}{+14.69}} & 95.37\textsuperscript{\textcolor{numpertdeltaimprove}{+10.52}} & 88.44\textsuperscript{\textcolor{numpertdeltaimprove}{+3.59}} \\
 & \mallperturbseventeen & 85.61 & 88.49\textsuperscript{\textcolor{numpertdeltaimprove}{+2.88}} & 78.24\textsuperscript{\textcolor{numpertdeltadecline}{-7.37}} & 81.05\textsuperscript{\textcolor{numpertdeltadecline}{-4.56}} & 98.08\textsuperscript{\textcolor{numpertdeltaimprove}{+12.47}} & 96.76\textsuperscript{\textcolor{numpertdeltaimprove}{+11.15}} & 93.52\textsuperscript{\textcolor{numpertdeltaimprove}{+7.91}} & 88.24\textsuperscript{\textcolor{numpertdeltaimprove}{+2.63}} \\
 & \mallperturbfour & 87.88 & 97.12\textsuperscript{\textcolor{numpertdeltaimprove}{+9.24}} & 95.37\textsuperscript{\textcolor{numpertdeltaimprove}{+7.49}} & 96.84\textsuperscript{\textcolor{numpertdeltaimprove}{+8.96}} & \textbf{100.00}\textsuperscript{\textcolor{numpertdeltaimprove}{+12.12}} & \textbf{100.00}\textsuperscript{\textcolor{numpertdeltaimprove}{+12.12}} & 97.69\textsuperscript{\textcolor{numpertdeltaimprove}{+9.81}} & 97.57\textsuperscript{\textcolor{numpertdeltaimprove}{+9.69}} \\
 & \mallperturb & 93.18 & 95.68\textsuperscript{\textcolor{numpertdeltaimprove}{+2.50}} & 96.30\textsuperscript{\textcolor{numpertdeltaimprove}{+3.12}} & 95.79\textsuperscript{\textcolor{numpertdeltaimprove}{+2.61}} & 98.08\textsuperscript{\textcolor{numpertdeltaimprove}{+4.90}} & \textbf{100.00}\textsuperscript{\textcolor{numpertdeltaimprove}{+6.82}} & \textbf{99.54}\textsuperscript{\textcolor{numpertdeltaimprove}{+6.36}} & 97.67\textsuperscript{\textcolor{numpertdeltaimprove}{+4.49}} \\
\midrule
\Block{6-1}{\rotate \groupthreetable} & \mapproximation & 91.67 & 96.40\textsuperscript{\textcolor{numpertdeltaimprove}{+4.73}} & 31.48\textsuperscript{\textcolor{numpertdeltadecline}{-60.19}} & 83.16\textsuperscript{\textcolor{numpertdeltadecline}{-8.51}} & 9.62\textsuperscript{\textcolor{numpertdeltadecline}{-82.05}} & 28.24\textsuperscript{\textcolor{numpertdeltadecline}{-63.43}} & 90.74\textsuperscript{\textcolor{numpertdeltadecline}{-0.93}} & 60.45\textsuperscript{\textcolor{numpertdeltadecline}{-31.22}} \\
 & \mnumeration & 95.45 & 93.53\textsuperscript{\textcolor{numpertdeltadecline}{-1.92}} & 97.22\textsuperscript{\textcolor{numpertdeltaimprove}{+1.77}} & 95.79\textsuperscript{\textcolor{numpertdeltaimprove}{+0.34}} & 7.69\textsuperscript{\textcolor{numpertdeltadecline}{-87.76}} & 21.76\textsuperscript{\textcolor{numpertdeltadecline}{-73.69}} & 96.76\textsuperscript{\textcolor{numpertdeltaimprove}{+1.31}} & 76.00\textsuperscript{\textcolor{numpertdeltadecline}{-19.45}} \\
 & \mrange & 93.94 & 67.63\textsuperscript{\textcolor{numpertdeltadecline}{-26.31}} & 41.20\textsuperscript{\textcolor{numpertdeltadecline}{-52.74}} & 97.37\textsuperscript{\textcolor{numpertdeltaimprove}{+3.43}} & 7.69\textsuperscript{\textcolor{numpertdeltadecline}{-86.25}} & 10.19\textsuperscript{\textcolor{numpertdeltadecline}{-83.75}} & 80.09\textsuperscript{\textcolor{numpertdeltadecline}{-13.85}} & 55.10\textsuperscript{\textcolor{numpertdeltadecline}{-38.84}} \\
 & \mnegation & 93.18 & 46.04\textsuperscript{\textcolor{numpertdeltadecline}{-47.14}} & 29.17\textsuperscript{\textcolor{numpertdeltadecline}{-64.01}} & 45.79\textsuperscript{\textcolor{numpertdeltadecline}{-47.39}} & 98.08\textsuperscript{\textcolor{numpertdeltaimprove}{+4.90}} & 17.13\textsuperscript{\textcolor{numpertdeltadecline}{-76.05}} & 59.26\textsuperscript{\textcolor{numpertdeltadecline}{-33.92}} & 41.79\textsuperscript{\textcolor{numpertdeltadecline}{-51.39}} \\
 & \mmasked & 93.56 & 24.46\textsuperscript{\textcolor{numpertdeltadecline}{-69.10}} & 17.59\textsuperscript{\textcolor{numpertdeltadecline}{-75.97}} & 26.32\textsuperscript{\textcolor{numpertdeltadecline}{-67.24}} & 26.92\textsuperscript{\textcolor{numpertdeltadecline}{-66.64}} & \textbf{100.00}\textsuperscript{\textcolor{numpertdeltaimprove}{+6.44}} & 39.35\textsuperscript{\textcolor{numpertdeltadecline}{-54.21}} & 42.47\textsuperscript{\textcolor{numpertdeltadecline}{-51.09}} \\
 & \mrandom & 94.70 & 95.68\textsuperscript{\textcolor{numpertdeltaimprove}{+0.98}} & 76.85\textsuperscript{\textcolor{numpertdeltadecline}{-17.85}} & 93.68\textsuperscript{\textcolor{numpertdeltadecline}{-1.02}} & 26.92\textsuperscript{\textcolor{numpertdeltadecline}{-67.78}} & 58.80\textsuperscript{\textcolor{numpertdeltadecline}{-35.90}} & 97.69\textsuperscript{\textcolor{numpertdeltaimprove}{+2.99}} & 80.56\textsuperscript{\textcolor{numpertdeltadecline}{-14.14}} \\
\midrule
\Block{7-1}{\rotate \groupfourtable} & \mnooriginal & 94.70 & 97.84\textsuperscript{\textcolor{numpertdeltaimprove}{+3.14}} & \textbf{98.61}\textsuperscript{\textcolor{numpertdeltaimprove}{+3.91}} & \textbf{98.42}\textsuperscript{\textcolor{numpertdeltaimprove}{+3.72}} & \textbf{100.00}\textsuperscript{\textcolor{numpertdeltaimprove}{+5.30}} & 99.54\textsuperscript{\textcolor{numpertdeltaimprove}{+4.84}} & 98.61\textsuperscript{\textcolor{numpertdeltaimprove}{+3.91}} & \textbf{98.74}\textsuperscript{\textcolor{numpertdeltaimprove}{+4.04}} \\
 & \mnoapproximation & 91.67 & 97.84\textsuperscript{\textcolor{numpertdeltaimprove}{+6.17}} & 97.69\textsuperscript{\textcolor{numpertdeltaimprove}{+6.02}} & 96.84\textsuperscript{\textcolor{numpertdeltaimprove}{+5.17}} & \textbf{100.00}\textsuperscript{\textcolor{numpertdeltaimprove}{+8.33}} & \textbf{100.00}\textsuperscript{\textcolor{numpertdeltaimprove}{+8.33}} & \textbf{99.54}\textsuperscript{\textcolor{numpertdeltaimprove}{+7.87}} & 98.54\textsuperscript{\textcolor{numpertdeltaimprove}{+6.87}} \\
 & \mnonumeration & 93.56 & 97.84\textsuperscript{\textcolor{numpertdeltaimprove}{+4.28}} & 62.04\textsuperscript{\textcolor{numpertdeltadecline}{-31.52}} & 95.79\textsuperscript{\textcolor{numpertdeltaimprove}{+2.23}} & \textbf{100.00}\textsuperscript{\textcolor{numpertdeltaimprove}{+6.44}} & \textbf{100.00}\textsuperscript{\textcolor{numpertdeltaimprove}{+6.44}} & 98.61\textsuperscript{\textcolor{numpertdeltaimprove}{+5.05}} & 90.67\textsuperscript{\textcolor{numpertdeltadecline}{-2.89}} \\
 & \mnorange & \textbf{96.21} & 95.68\textsuperscript{\textcolor{numpertdeltadecline}{-0.53}} & 93.06\textsuperscript{\textcolor{numpertdeltadecline}{-3.15}} & 95.26\textsuperscript{\textcolor{numpertdeltadecline}{-0.95}} & 98.08\textsuperscript{\textcolor{numpertdeltaimprove}{+1.87}} & 99.54\textsuperscript{\textcolor{numpertdeltaimprove}{+3.33}} & 97.69\textsuperscript{\textcolor{numpertdeltaimprove}{+1.48}} & 96.41\textsuperscript{\textcolor{numpertdeltaimprove}{+0.20}} \\
 & \mnonegation & 91.67 & \textbf{98.56}\textsuperscript{\textcolor{numpertdeltaimprove}{+6.89}} & 98.15\textsuperscript{\textcolor{numpertdeltaimprove}{+6.48}} & 96.84\textsuperscript{\textcolor{numpertdeltaimprove}{+5.17}} & 67.31\textsuperscript{\textcolor{numpertdeltadecline}{-24.36}} & \textbf{100.00}\textsuperscript{\textcolor{numpertdeltaimprove}{+8.33}} & 97.69\textsuperscript{\textcolor{numpertdeltaimprove}{+6.02}} & 96.70\textsuperscript{\textcolor{numpertdeltaimprove}{+5.03}} \\
 & \mnomasked & 93.56 & 95.68\textsuperscript{\textcolor{numpertdeltaimprove}{+2.12}} & 95.83\textsuperscript{\textcolor{numpertdeltaimprove}{+2.27}} & 96.84\textsuperscript{\textcolor{numpertdeltaimprove}{+3.28}} & 98.08\textsuperscript{\textcolor{numpertdeltaimprove}{+4.52}} & 13.43\textsuperscript{\textcolor{numpertdeltadecline}{-80.13}} & 97.22\textsuperscript{\textcolor{numpertdeltaimprove}{+3.66}} & 79.10\textsuperscript{\textcolor{numpertdeltadecline}{-14.46}} \\
 & \mnorandom & 95.83 & 94.24\textsuperscript{\textcolor{numpertdeltadecline}{-1.59}} & 94.91\textsuperscript{\textcolor{numpertdeltadecline}{-0.92}} & 94.21\textsuperscript{\textcolor{numpertdeltadecline}{-1.62}} & 98.08\textsuperscript{\textcolor{numpertdeltaimprove}{+2.25}} & 99.54\textsuperscript{\textcolor{numpertdeltaimprove}{+3.71}} & 93.98\textsuperscript{\textcolor{numpertdeltadecline}{-1.85}} & 95.63\textsuperscript{\textcolor{numpertdeltadecline}{-0.20}} \\
\bottomrule
\end{NiceTabular}
}
\caption{\footnotesize \ttf\ accuracy (\%) on unperturbed claims (\textbf{Orig.}, left of dotted rule) and label-flipping perturbations. Deltas vs.\ \textbf{Orig.} within each row. M.\ Avg.\ is the micro-average of accuracy across perturbation types.}
\label{tab:ttf}
\end{table*}

%% file: emnlp2026-numpert-finetune-05-revised.tex
\section{Results}
\label{sec:results}

\input{tables/results_q4_k_m/quantemp_results/F-_F}

\input{tables/results_q4_k_m/quantemp_results/T-_T}
\input{tables/results_q4_k_m/generalization}

Tables~\ref{tab:ttf}--\ref{tab:ttt} are organised into four model groups: \groupone\ (baselines and clean fine-tune), \grouptwo\ (all-perturbation training), \groupthree\ (single-type training), and \groupfour\ (leave-one-out training).

\subsection{\texttt{True} $\rightarrow$  \texttt{False}}
\label{sec:ttf}

The \ttf\ evaluation is the most diagnostic setting: numerical values are perturbed to flip the label, so a model that merely anchors on surface form will be confidently wrong.

The \groupone\ models' accuracy collapses for the \negative\ and \masked\ categories, with \mnoperturb\ dropping by 83--90\% on perturbed examples. 


QLoRA models in \grouptwo\ are far more robust than \groupone, reaching near-perfect accuracy on \negative\ and \masked. Unlike \groupthree, which collapses on unseen types, \grouptwo\ shows model size steadily improving robustness, with \mallperturb\ reaching mid-to-high-90s accuracy and the smallest fluctuations.

The leave-one-out group (\groupfour) has the best accuracies overall, but \mnonumeration, \mnonegation, and especially \mnomasked\ collapse on their omitted type ($-$31.52\,pp, $-$24.36\,pp, $-$80.13\,pp). The most capable models are \mnooriginal\ and \mnorange, with the smallest fluctuations across types.
\masked\ detection does not transfer from any other training type; it must be trained directly. Excluding \numeration\ hurts more than zero-shot, while excluding \approximation, \range, or \random\ has little negative effect (see Section~\ref{sec:discussion}).

\subsection{\texttt{True} $\rightarrow$ \texttt{True}}
\label{sec:ttt}
Table~\ref{tab:ttt} shows results for \ttt, excluding \negative, \random, and \masked\ as they produce only non-structure-preserving perturbations.

The \groupone's and the \grouptwo's models generally show the smallest accuracy fluctuations between original and perturbed examples. The \mnoperturb\ model achieves mid-90s accuracy ($\pm0.57\%$).

The \groupthree\ and \groupfour\ models fluctuate more: single-type training provides no immunity to other perturbation types (e.g.\ \mnumeration\ drops $-$61.24\% on \range), and leave-one-out training exposes similar gaps (e.g.\ \mnorange\ on \range, \mnoapproximation\ on \approximation). 
Adversarial fine-tuning is decisively beneficial: even \mallperturbsix\ exceeds \mbasethirtytwo\ across all perturbation types, and excluding \random\ from training (\mnorandom) produces the strongest overall profile. The full analysis is in Section~\ref{sec:discussion}.

\subsection{\texttt{False} $\rightarrow$ \texttt{False}}
\label{sec:fff}

In the \ftf\ evaluation (Table~\ref{tab:ftf}), models are robustly accurate, in the mid-90s across types. Fine-tuning on \negative, \masked, and \random\ saturates the small \groupone\ blind spots on those types. \mrandom\ is nearly perfect (99.53\%) on \range, where most other models struggle most.
All models above 0.6B are robustly accurate on false claims; the main exceptions are \range, which causes accuracy drops across groups, and \approximation, where leave-one-out training counterintuitively yields the best accuracy (99.23\%).

\subsection{Cross-Dataset Generalisation}
\label{sec:generalization}
Table~\ref{tab:generalization_results} shows how \mbase, \mnoapproximation, and \mallperturb\ transfer to two unseen datasets. We restrict cross-dataset evaluation to baseline and all-perturbs models; other configurations generalised comparably and are omitted for brevity.

\paragraph{CLEF 2025 Spanish.}
Both fine-tuned models beat \mbase\ overall (97.48\% and 97.36\% vs.\ 96.44\%), with the largest gains on value-changing (T2) and obfuscation (T3) perturbations (99.44--99.54\%, $+$2\,pp).
That aggregate understates the effect considerably. The split is 95.8\%
\texttt{False}, and the transition groups pool the instances whose label
was flipped with the far larger set that was already \texttt{False}: the
value-changing (T2) column combines 553 \ttf\ instances with 5{,}578 \ftf\ ones, so it
tracks a majority on which every model is already near ceiling.
Measured \emph{within} transitions, \mallperturb\ improves \ttf\
accuracy from 73.71\% to 97.07\% ($+$23.36\,pp), a gain an order of
magnitude larger than the aggregate suggests, and every other perturbation
type improves by 13--25\,pp. The largest gain by far is on \negative\
($+$64.71\,pp), where \mbase\ reaches only 35.29\% -- below chance --
against 100.00\% for both fine-tuned models. The gain is confined to the flipped
cases: \ttt\ is essentially unchanged (79.04\% to 79.55\%) and \ftf\ is
already near ceiling (97.44\% to 98.28\%), so fine-tuning does not make
the model better at recognising claims that stay true, it stops it
confirming claims whose quantity was altered.
Appendix~\ref{app:spanish_perturbation} gives the full per-type
breakdown, and notes that \mbase\ leaves 75.6\% of \ttf\ instances
unparseable, so its 73.71\% rests on 175 of 717 instances.

\paragraph{AVeriTeC 2.0 dev.}
On parseable responses the baseline reaches 95.31\% against 93.20\% for \mallperturb\ ($-$2.11\,pp), a deficit confined to T1; the two are within 0.4\,pp on T2 and T3 (Table~\ref{tab:generalization_results}). It is not significant once paired ($p = 0.511$, Section~\ref{sec:significance}) and rests on very different denominators: \mbase\ returns no parseable verdict on 281 of 750 instances (37.5\%), against zero for the fine-tuned models. On AVeriTeC the advantage is coverage, not accuracy.

\subsection{Statistical Significance}
\label{sec:significance}

The comparisons so far are single-run point estimates. We recompute the cross-dataset results from the stored per-instance predictions with a paired bootstrap (10{,}000 resamples) and an exact McNemar test. Because \mbase\ leaves many instances unparseable while the fine-tuned models leave none, we report three protocols (Appendix~\ref{app:significance}). On CLEF Spanish the gain is significant under every one: paired over the 9{,}090 instances both models parse, \mallperturb\ gains $+$2.33\,pp ($p = 8.0\mathrm{e}{-}41$) in every perturbation direction. Since that split is 95.8\% \texttt{False}, macro-F1 is more informative: it rises 8.43\,pp, almost entirely in the minority \texttt{True} class, whose F1 rises 16.4\,pp from a baseline precision of just 42.4\% (Table~\ref{tab:spanish_per_class}). On AVeriTeC the apparent parseable-only deficit is \emph{not} significant once paired ($-$1.07\,pp, $p = 0.511$); the real advantage there is coverage, as \mbase\ returns no parseable verdict on 37.5\% of instances against zero for \mallperturb. We therefore describe AVeriTeC as a coverage rather than an accuracy gain.

\subsection{Sensitivity to Numerical Distance}
\label{sec:distance}

Our taxonomy only coarsely distinguishes a perturbation that nudges a
value from one that replaces it with an implausible figure. Binning each
perturbed instance by relative numerical distance separates the two
(Appendix~\ref{app:distance}). The benefit of adversarial fine-tuning is
largest exactly on the near-misses: $+$7.50\,pp within 10\% relative
distance and $+$8.44\,pp between 10\% and 25\%, falling monotonically to
$+$2.91\,pp beyond 100\%, every bin individually significant. This bears
directly on the concern that the models might be recognising
perturbation artefacts rather than comparing quantities: an artefact
detector would improve uniformly, or most where the edit is most
conspicuous, whereas the observed profile is the opposite.

\subsection{Evidence-Side Numerical Revisions}
\label{sec:vitaminc}
\input{tables/results_q4_k_m/vitaminc_crossdataset}

The perturbations studied so far all edit the \emph{claim} while holding
the evidence fixed, leaving open whether that robustness is
direction-specific. We therefore evaluate on VitaminC
\citep{Schuster:2021:NAACL}, which pairs each claim with two snapshots
of one Wikipedia sentence, before and after a factual revision: the claim is unaltered natural text, and the quantity
deciding the verdict changes in the \emph{evidence}. Restricted to real,
genuinely numerical revisions, it yields 11{,}887 test instances, including
5{,}227 \emph{contrastive pairs} differing solely by a number
in the evidence (Appendix~\ref{app:vitaminc_details}); we therefore also
report \emph{pair accuracy}, the share of pairs with \emph{both}
snapshots correct.

\paragraph{Claim-side robustness does not transfer.}
\mallperturb\ reaches only 78.35\% accuracy and 54.45\% pair accuracy,
\emph{below} the untuned \mbase\ at 90.96\% and 84.25\% ($-$12.60\,pp
and $-$29.81\,pp, exact McNemar $p < 10^{-190}$;
Table~\ref{tab:vitaminc_main}). The errors are
strongly asymmetric: \mallperturb\ wrongly confirms a refuting snapshot
on 31.5\% of pairs but wrongly refutes a supporting one on only 15.0\%,
whereas \mbase\ is balanced (8.4\% and 8.8\%). Inspecting the 1,426
pairs where only \mallperturb\ errs shows a consistent pattern:
``\$24.9 million'' is judged to support ``grossed more than \$25
million''. The model checks whether the two quantities are approximately
\emph{equal} rather than evaluating the inequality the claim asserts:
sufficient on our claim-side perturbations, which move the value away
from the evidence, but not on bounded claims. Prompting does not repair it: \scotfull\ (\scot) and \pap\ are
neutral or harmful, and a \tcpfull\ (\tcp) naming the failure directly
recovers only about half the deficit
(Appendix~\ref{app:vitaminc_results}).

\paragraph{Training on evidence-side revisions closes it, and the two competencies are complementary.}
Trained with the same recipe on VitaminC itself, \mvitaminc\ reaches 98.05\%
accuracy and 96.56\% pair accuracy, exceeding \mbase\ by $+$7.09\,pp and
$+$12.30\,pp ($p < 10^{-140}$), and wrong confirmations fall from 31.5\%
to 2.5\%. The recipe transfers even though the perturbation direction,
the discrepancy's source and the evidence length all differ; the
resulting \emph{model} does not. Specialisation is costly in both
directions (Table~\ref{tab:vitaminc_crossdataset}): \mvitaminc\ scores
only 80.86\% on QuanTemp against \mallperturb's 97.52\%, below even
\mbase. Training on both corpora jointly incurs no trade-off: \mmixed\
attains
98.25\% accuracy on VitaminC and 98.20\% on QuanTemp, and is
significantly better than \emph{each specialist on that specialist's own
test set} ($p = 1.0\mathrm{e}{-4}$ and $p = 0.002$). Claim-side and
evidence-side numerical robustness are thus distinct competencies,
neither obtainable from the other, but jointly learnable.

%% file: tables/results_q4_k_m/quantemp_results/F-_F.tex
\begin{table*}[ht]
\centering
\definecolor{numpertdeltaimprove}{RGB}{0,100,0}
\definecolor{numpertdeltadecline}{RGB}{180,35,24}
\small
\setlength{\tabcolsep}{3pt}
\resizebox{\textwidth}{!}{%
\begin{NiceTabular}{l|cccccccc}[first-col]
\toprule
 & \textbf{Model} & \textbf{Orig.} & \textbf{Approx.} & \textbf{Num.} & \textbf{Range} & \textbf{Neg.} & \textbf{Mask.} & \textbf{Rand.} & \textbf{M. Avg.} \\
\midrule
\Block{6-1}{\rotate \grouponetable} & \mgptfive & 95.33 & 95.84\textsuperscript{\textcolor{numpertdeltaimprove}{+0.51}} & 96.51\textsuperscript{\textcolor{numpertdeltaimprove}{+1.18}} & 94.65\textsuperscript{\textcolor{numpertdeltadecline}{-0.68}} & 92.22\textsuperscript{\textcolor{numpertdeltadecline}{-3.11}} & 96.69\textsuperscript{\textcolor{numpertdeltaimprove}{+1.36}} & 97.75\textsuperscript{\textcolor{numpertdeltaimprove}{+2.42}} & 96.03\textsuperscript{\textcolor{numpertdeltaimprove}{+0.70}} \\
& \mgemini & 91.48 & 90.77\textsuperscript{\textcolor{numpertdeltadecline}{-0.71}} & 90.87\textsuperscript{\textcolor{numpertdeltadecline}{-0.61}} & 89.61\textsuperscript{\textcolor{numpertdeltadecline}{-1.87}} & 90.00\textsuperscript{\textcolor{numpertdeltadecline}{-1.48}} & 88.55\textsuperscript{\textcolor{numpertdeltadecline}{-2.93}} & 96.77\textsuperscript{\textcolor{numpertdeltaimprove}{+5.29}} & 91.35\textsuperscript{\textcolor{numpertdeltadecline}{-0.13}} \\
& \mbase & 94.30 & 95.51\textsuperscript{\textcolor{numpertdeltaimprove}{+1.21}} & 94.65\textsuperscript{\textcolor{numpertdeltaimprove}{+0.35}} & 92.18\textsuperscript{\textcolor{numpertdeltadecline}{-2.12}} & 93.75\textsuperscript{\textcolor{numpertdeltadecline}{-0.55}} & 94.03\textsuperscript{\textcolor{numpertdeltadecline}{-0.27}} & 97.70\textsuperscript{\textcolor{numpertdeltaimprove}{+3.40}} & 94.83\textsuperscript{\textcolor{numpertdeltaimprove}{+0.53}} \\
 & \mbasetwoshot & 95.42 & 96.43\textsuperscript{\textcolor{numpertdeltaimprove}{+1.01}} & 95.99\textsuperscript{\textcolor{numpertdeltaimprove}{+0.57}} & 95.23\textsuperscript{\textcolor{numpertdeltadecline}{-0.19}} & 94.51\textsuperscript{\textcolor{numpertdeltadecline}{-0.91}} & 94.37\textsuperscript{\textcolor{numpertdeltadecline}{-1.05}} & 98.60\textsuperscript{\textcolor{numpertdeltaimprove}{+3.18}} & 96.08\textsuperscript{\textcolor{numpertdeltaimprove}{+0.66}} \\
 & \mbasethirtytwo & 95.24 & 96.64\textsuperscript{\textcolor{numpertdeltaimprove}{+1.40}} & 95.37\textsuperscript{\textcolor{numpertdeltaimprove}{+0.13}} & 94.67\textsuperscript{\textcolor{numpertdeltadecline}{-0.57}} & 93.33\textsuperscript{\textcolor{numpertdeltadecline}{-1.91}} & 95.77\textsuperscript{\textcolor{numpertdeltaimprove}{+0.53}} & 97.77\textsuperscript{\textcolor{numpertdeltaimprove}{+2.53}} & 95.96\textsuperscript{\textcolor{numpertdeltaimprove}{+0.72}} \\
 & \mnoperturb & 96.75 & 97.70\textsuperscript{\textcolor{numpertdeltaimprove}{+0.95}} & 96.41\textsuperscript{\textcolor{numpertdeltadecline}{-0.34}} & 96.92\textsuperscript{\textcolor{numpertdeltaimprove}{+0.17}} & 96.70\textsuperscript{\textcolor{numpertdeltadecline}{-0.05}} & 96.61\textsuperscript{\textcolor{numpertdeltadecline}{-0.14}} & 98.20\textsuperscript{\textcolor{numpertdeltaimprove}{+1.45}} & 97.13\textsuperscript{\textcolor{numpertdeltaimprove}{+0.38}} \\
\midrule
\Block{4-1}{\rotate \grouptwotable} & \mptfulldata & 94.65 & 96.84\textsuperscript{\textcolor{numpertdeltaimprove}{+2.19}} & 95.68\textsuperscript{\textcolor{numpertdeltaimprove}{+1.03}} & 97.31\textsuperscript{\textcolor{numpertdeltaimprove}{+2.66}} & 95.24\textsuperscript{\textcolor{numpertdeltaimprove}{+0.59}} & 95.69\textsuperscript{\textcolor{numpertdeltaimprove}{+1.04}} & 97.33\textsuperscript{\textcolor{numpertdeltaimprove}{+2.68}} & 96.35\textsuperscript{\textcolor{numpertdeltaimprove}{+1.70}} \\
 & \mallperturbsix & 95.12 & 94.90\textsuperscript{\textcolor{numpertdeltadecline}{-0.22}} & 94.41\textsuperscript{\textcolor{numpertdeltadecline}{-0.71}} & 84.60\textsuperscript{\textcolor{numpertdeltadecline}{-10.52}} & \textbf{100.00}\textsuperscript{\textcolor{numpertdeltaimprove}{+4.88}} & \textbf{100.00}\textsuperscript{\textcolor{numpertdeltaimprove}{+4.88}} & 99.80\textsuperscript{\textcolor{numpertdeltaimprove}{+4.68}} & 95.27\textsuperscript{\textcolor{numpertdeltaimprove}{+0.15}} \\
 & \mallperturbseventeen & 96.10 & 94.64\textsuperscript{\textcolor{numpertdeltadecline}{-1.46}} & 95.01\textsuperscript{\textcolor{numpertdeltadecline}{-1.09}} & 92.18\textsuperscript{\textcolor{numpertdeltadecline}{-3.92}} & \textbf{100.00}\textsuperscript{\textcolor{numpertdeltaimprove}{+3.90}} & 99.80\textsuperscript{\textcolor{numpertdeltaimprove}{+3.70}} & 99.40\textsuperscript{\textcolor{numpertdeltaimprove}{+3.30}} & 96.55\textsuperscript{\textcolor{numpertdeltaimprove}{+0.45}} \\
 & \mallperturbfour & \textbf{98.37} & 97.45\textsuperscript{\textcolor{numpertdeltadecline}{-0.92}} & \textbf{97.80}\textsuperscript{\textcolor{numpertdeltadecline}{-0.57}} & 93.60\textsuperscript{\textcolor{numpertdeltadecline}{-4.77}} & \textbf{100.00}\textsuperscript{\textcolor{numpertdeltaimprove}{+1.63}} & \textbf{100.00}\textsuperscript{\textcolor{numpertdeltaimprove}{+1.63}} & \textbf{100.00}\textsuperscript{\textcolor{numpertdeltaimprove}{+1.63}} & 98.01\textsuperscript{\textcolor{numpertdeltadecline}{-0.36}} \\
 & \mallperturb & 97.40 & 96.94\textsuperscript{\textcolor{numpertdeltadecline}{-0.46}} & 97.21\textsuperscript{\textcolor{numpertdeltadecline}{-0.19}} & 95.02\textsuperscript{\textcolor{numpertdeltadecline}{-2.38}} & \textbf{100.00}\textsuperscript{\textcolor{numpertdeltaimprove}{+2.60}} & \textbf{100.00}\textsuperscript{\textcolor{numpertdeltaimprove}{+2.60}} & \textbf{100.00}\textsuperscript{\textcolor{numpertdeltaimprove}{+2.60}} & 98.05\textsuperscript{\textcolor{numpertdeltaimprove}{+0.65}} \\
\midrule
\Block{6-1}{\rotate \groupthreetable} & \mapproximation & 96.42 & 95.15\textsuperscript{\textcolor{numpertdeltadecline}{-1.27}} & 96.61\textsuperscript{\textcolor{numpertdeltaimprove}{+0.19}} & 96.45\textsuperscript{\textcolor{numpertdeltaimprove}{+0.03}} & 95.60\textsuperscript{\textcolor{numpertdeltadecline}{-0.82}} & 97.01\textsuperscript{\textcolor{numpertdeltaimprove}{+0.59}} & 99.00\textsuperscript{\textcolor{numpertdeltaimprove}{+2.58}} & 96.89\textsuperscript{\textcolor{numpertdeltaimprove}{+0.47}} \\
 & \mnumeration & 96.26 & 98.21\textsuperscript{\textcolor{numpertdeltaimprove}{+1.95}} & 96.61\textsuperscript{\textcolor{numpertdeltaimprove}{+0.35}} & 98.82\textsuperscript{\textcolor{numpertdeltaimprove}{+2.56}} & 94.51\textsuperscript{\textcolor{numpertdeltadecline}{-1.75}} & 96.21\textsuperscript{\textcolor{numpertdeltadecline}{-0.05}} & \textbf{100.00}\textsuperscript{\textcolor{numpertdeltaimprove}{+3.74}} & 97.80\textsuperscript{\textcolor{numpertdeltaimprove}{+1.54}} \\
 & \mrange & 96.91 & 96.94\textsuperscript{\textcolor{numpertdeltaimprove}{+0.03}} & 96.61\textsuperscript{\textcolor{numpertdeltadecline}{-0.30}} & 94.08\textsuperscript{\textcolor{numpertdeltadecline}{-2.83}} & 96.70\textsuperscript{\textcolor{numpertdeltadecline}{-0.21}} & 95.61\textsuperscript{\textcolor{numpertdeltadecline}{-1.30}} & 99.20\textsuperscript{\textcolor{numpertdeltaimprove}{+2.29}} & 96.55\textsuperscript{\textcolor{numpertdeltadecline}{-0.36}} \\
 & \mnegation & 95.12 & 96.94\textsuperscript{\textcolor{numpertdeltaimprove}{+1.82}} & 95.41\textsuperscript{\textcolor{numpertdeltaimprove}{+0.29}} & 96.68\textsuperscript{\textcolor{numpertdeltaimprove}{+1.56}} & \textbf{100.00}\textsuperscript{\textcolor{numpertdeltaimprove}{+4.88}} & 95.81\textsuperscript{\textcolor{numpertdeltaimprove}{+0.69}} & 98.80\textsuperscript{\textcolor{numpertdeltaimprove}{+3.68}} & 96.84\textsuperscript{\textcolor{numpertdeltaimprove}{+1.72}} \\
 & \mmasked & 96.91 & 97.45\textsuperscript{\textcolor{numpertdeltaimprove}{+0.54}} & 96.61\textsuperscript{\textcolor{numpertdeltadecline}{-0.30}} & 96.45\textsuperscript{\textcolor{numpertdeltadecline}{-0.46}} & 96.70\textsuperscript{\textcolor{numpertdeltadecline}{-0.21}} & \textbf{100.00}\textsuperscript{\textcolor{numpertdeltaimprove}{+3.09}} & 98.60\textsuperscript{\textcolor{numpertdeltaimprove}{+1.69}} & 97.84\textsuperscript{\textcolor{numpertdeltaimprove}{+0.93}} \\
 & \mrandom & 94.96 & 97.96\textsuperscript{\textcolor{numpertdeltaimprove}{+3.00}} & 95.41\textsuperscript{\textcolor{numpertdeltaimprove}{+0.45}} & \textbf{99.53}\textsuperscript{\textcolor{numpertdeltaimprove}{+4.57}} & 97.80\textsuperscript{\textcolor{numpertdeltaimprove}{+2.84}} & 98.00\textsuperscript{\textcolor{numpertdeltaimprove}{+3.04}} & \textbf{100.00}\textsuperscript{\textcolor{numpertdeltaimprove}{+5.04}} & 98.13\textsuperscript{\textcolor{numpertdeltaimprove}{+3.17}} \\
\midrule
\Block{7-1}{\rotate \groupfourtable} & \mnooriginal & 95.45 & 94.39\textsuperscript{\textcolor{numpertdeltadecline}{-1.06}} & 96.01\textsuperscript{\textcolor{numpertdeltaimprove}{+0.56}} & 92.18\textsuperscript{\textcolor{numpertdeltadecline}{-3.27}} & \textbf{100.00}\textsuperscript{\textcolor{numpertdeltaimprove}{+4.55}} & \textbf{100.00}\textsuperscript{\textcolor{numpertdeltaimprove}{+4.55}} & 99.80\textsuperscript{\textcolor{numpertdeltaimprove}{+4.35}} & 96.84\textsuperscript{\textcolor{numpertdeltaimprove}{+1.39}} \\
 & \mnoapproximation & 97.07 & \textbf{99.23}\textsuperscript{\textcolor{numpertdeltaimprove}{+2.16}} & 97.01\textsuperscript{\textcolor{numpertdeltadecline}{-0.06}} & 93.13\textsuperscript{\textcolor{numpertdeltadecline}{-3.94}} & \textbf{100.00}\textsuperscript{\textcolor{numpertdeltaimprove}{+2.93}} & \textbf{100.00}\textsuperscript{\textcolor{numpertdeltaimprove}{+2.93}} & \textbf{100.00}\textsuperscript{\textcolor{numpertdeltaimprove}{+2.93}} & 98.05\textsuperscript{\textcolor{numpertdeltaimprove}{+0.98}} \\
 & \mnonumeration & 96.75 & 96.43\textsuperscript{\textcolor{numpertdeltadecline}{-0.32}} & 96.21\textsuperscript{\textcolor{numpertdeltadecline}{-0.54}} & 91.47\textsuperscript{\textcolor{numpertdeltadecline}{-5.28}} & \textbf{100.00}\textsuperscript{\textcolor{numpertdeltaimprove}{+3.25}} & \textbf{100.00}\textsuperscript{\textcolor{numpertdeltaimprove}{+3.25}} & \textbf{100.00}\textsuperscript{\textcolor{numpertdeltaimprove}{+3.25}} & 97.14\textsuperscript{\textcolor{numpertdeltaimprove}{+0.39}} \\
 & \mnorange & 95.12 & 95.15\textsuperscript{\textcolor{numpertdeltaimprove}{+0.03}} & 94.61\textsuperscript{\textcolor{numpertdeltadecline}{-0.51}} & 95.50\textsuperscript{\textcolor{numpertdeltaimprove}{+0.38}} & \textbf{100.00}\textsuperscript{\textcolor{numpertdeltaimprove}{+4.88}} & 99.80\textsuperscript{\textcolor{numpertdeltaimprove}{+4.68}} & 99.60\textsuperscript{\textcolor{numpertdeltaimprove}{+4.48}} & 97.18\textsuperscript{\textcolor{numpertdeltaimprove}{+2.06}} \\
 & \mnonegation & 97.72 & 97.70\textsuperscript{\textcolor{numpertdeltadecline}{-0.02}} & \textbf{97.80}\textsuperscript{\textcolor{numpertdeltaimprove}{+0.08}} & 97.39\textsuperscript{\textcolor{numpertdeltadecline}{-0.33}} & 97.80\textsuperscript{\textcolor{numpertdeltaimprove}{+0.08}} & \textbf{100.00}\textsuperscript{\textcolor{numpertdeltaimprove}{+2.28}} & \textbf{100.00}\textsuperscript{\textcolor{numpertdeltaimprove}{+2.28}} & \textbf{98.63}\textsuperscript{\textcolor{numpertdeltaimprove}{+0.91}} \\
 & \mnomasked & 96.10 & 95.92\textsuperscript{\textcolor{numpertdeltadecline}{-0.18}} & 96.01\textsuperscript{\textcolor{numpertdeltadecline}{-0.09}} & 92.65\textsuperscript{\textcolor{numpertdeltadecline}{-3.45}} & \textbf{100.00}\textsuperscript{\textcolor{numpertdeltaimprove}{+3.90}} & 94.81\textsuperscript{\textcolor{numpertdeltadecline}{-1.29}} & \textbf{100.00}\textsuperscript{\textcolor{numpertdeltaimprove}{+3.90}} & 96.14\textsuperscript{\textcolor{numpertdeltaimprove}{+0.04}} \\
 & \mnorandom & 94.63 & 94.64\textsuperscript{\textcolor{numpertdeltaimprove}{+0.01}} & 94.61\textsuperscript{\textcolor{numpertdeltadecline}{-0.02}} & 91.94\textsuperscript{\textcolor{numpertdeltadecline}{-2.69}} & \textbf{100.00}\textsuperscript{\textcolor{numpertdeltaimprove}{+5.37}} & \textbf{100.00}\textsuperscript{\textcolor{numpertdeltaimprove}{+5.37}} & 99.80\textsuperscript{\textcolor{numpertdeltaimprove}{+5.17}} & 96.55\textsuperscript{\textcolor{numpertdeltaimprove}{+1.92}} \\
\bottomrule
\end{NiceTabular}
}
\caption{\footnotesize \ftf\ accuracy (\%) on unperturbed claims (\textbf{Orig.}) and \texttt{False}-labelled perturbations. Deltas vs.\ \textbf{Orig.} within each row; M.\ Avg.\ is the micro-average of accuracy across perturbation types.}
\label{tab:ftf}
\end{table*}

%% file: tables/results_q4_k_m/quantemp_results/T-_T.tex
\begin{table}[ht]
\centering
\definecolor{numpertdeltaimprove}{RGB}{0,100,0}
\definecolor{numpertdeltadecline}{RGB}{180,35,24}
\small
\setlength{\tabcolsep}{4pt}
\resizebox{\columnwidth}{!}{%
\begin{NiceTabular}{l|cccccc}[first-col]
\toprule
 & \textbf{Model} & \textbf{Approx.} & \textbf{Num.} & \textbf{Range} & \textbf{M. Avg.} \\
\midrule
\Block{6-1}{\rotate \grouponetable}  & \mgptfive & 77.04\textsuperscript{\textcolor{numpertdeltadecline}{-1.08}} & 81.43\textsuperscript{\textcolor{numpertdeltaimprove}{+3.31}} & 82.39\textsuperscript{\textcolor{numpertdeltaimprove}{+4.27}} & 80.28\textsuperscript{\textcolor{numpertdeltaimprove}{+2.16}} \\
& \mgemini & 70.80\textsuperscript{\textcolor{numpertdeltadecline}{-3.91}} & 74.65\textsuperscript{\textcolor{numpertdeltadecline}{-0.06}} & 74.33\textsuperscript{\textcolor{numpertdeltadecline}{-0.38}} & 73.56\textsuperscript{\textcolor{numpertdeltadecline}{-1.15}} \\
& \mbase & 80.51\textsuperscript{\textcolor{numpertdeltadecline}{-5.84}} & 87.44\textsuperscript{\textcolor{numpertdeltaimprove}{+1.09}} & 85.29\textsuperscript{\textcolor{numpertdeltadecline}{-1.06}} & 84.92\textsuperscript{\textcolor{numpertdeltadecline}{-1.43}} \\
 & \mbasetwoshot & 73.33\textsuperscript{\textcolor{numpertdeltadecline}{-8.63}} & 82.94\textsuperscript{\textcolor{numpertdeltaimprove}{+0.98}} & 79.56\textsuperscript{\textcolor{numpertdeltadecline}{-2.40}} & 79.31\textsuperscript{\textcolor{numpertdeltadecline}{-2.65}} \\
 & \mbasethirtytwo & 77.44\textsuperscript{\textcolor{numpertdeltadecline}{-8.11}} & 82.30\textsuperscript{\textcolor{numpertdeltadecline}{-3.25}} & 81.01\textsuperscript{\textcolor{numpertdeltadecline}{-4.54}} & 80.61\textsuperscript{\textcolor{numpertdeltadecline}{-4.94}} \\
 & \mnoperturb & 96.40\textsuperscript{\textcolor{numpertdeltaimprove}{+0.57}} & \textbf{95.83} & 95.26\textsuperscript{\textcolor{numpertdeltadecline}{-0.57}} & 95.78\textsuperscript{\textcolor{numpertdeltadecline}{-0.05}} \\
\midrule
\Block{4-1}{\rotate \grouptwotable} & \mptfulldata & 75.00\textsuperscript{\textcolor{numpertdeltadecline}{-6.71}} & 80.10\textsuperscript{\textcolor{numpertdeltadecline}{-1.61}} & 78.69\textsuperscript{\textcolor{numpertdeltadecline}{-3.02}} & 77.93\textsuperscript{\textcolor{numpertdeltadecline}{-3.78}} \\
& \mallperturbsix & 89.21\textsuperscript{\textcolor{numpertdeltaimprove}{+4.36}} & 82.87\textsuperscript{\textcolor{numpertdeltadecline}{-1.98}} & 88.95\textsuperscript{\textcolor{numpertdeltaimprove}{+4.10}} & 86.61\textsuperscript{\textcolor{numpertdeltaimprove}{+1.76}} \\
 & \mallperturbseventeen & 89.93\textsuperscript{\textcolor{numpertdeltaimprove}{+4.32}} & 88.89\textsuperscript{\textcolor{numpertdeltaimprove}{+3.28}} & 91.58\textsuperscript{\textcolor{numpertdeltaimprove}{+5.97}} & 90.09\textsuperscript{\textcolor{numpertdeltaimprove}{+4.48}} \\
 & \mallperturbfour & 89.93\textsuperscript{\textcolor{numpertdeltaimprove}{+2.05}} & 89.35\textsuperscript{\textcolor{numpertdeltaimprove}{+1.47}} & 91.05\textsuperscript{\textcolor{numpertdeltaimprove}{+3.17}} & 90.09\textsuperscript{\textcolor{numpertdeltaimprove}{+2.21}} \\
 & \mallperturb & 89.21\textsuperscript{\textcolor{numpertdeltadecline}{-3.97}} & 91.20\textsuperscript{\textcolor{numpertdeltadecline}{-1.98}} & 93.16\textsuperscript{\textcolor{numpertdeltadecline}{-0.02}} & 91.38\textsuperscript{\textcolor{numpertdeltadecline}{-1.80}} \\
\midrule
\Block{6-1}{\rotate \groupthreetable} & \mapproximation  & 93.53\textsuperscript{\textcolor{numpertdeltaimprove}{+1.86}} & 93.98\textsuperscript{\textcolor{numpertdeltaimprove}{+2.31}} & 82.11\textsuperscript{\textcolor{numpertdeltadecline}{-9.56}} & 89.73\textsuperscript{\textcolor{numpertdeltadecline}{-1.94}} \\
 & \mnumeration  & 71.94\textsuperscript{\textcolor{numpertdeltadecline}{-23.51}} & 94.91\textsuperscript{\textcolor{numpertdeltadecline}{-0.54}} & 34.21\textsuperscript{\textcolor{numpertdeltadecline}{-61.24}} & 67.89\textsuperscript{\textcolor{numpertdeltadecline}{-27.56}} \\
 & \mrange  & 95.68\textsuperscript{\textcolor{numpertdeltaimprove}{+1.74}} & 93.98\textsuperscript{\textcolor{numpertdeltaimprove}{+0.04}} & 95.26\textsuperscript{\textcolor{numpertdeltaimprove}{+1.32}} & 94.86\textsuperscript{\textcolor{numpertdeltaimprove}{+0.92}} \\
 & \mnegation  & 88.49\textsuperscript{\textcolor{numpertdeltadecline}{-4.69}} & 92.59\textsuperscript{\textcolor{numpertdeltadecline}{-0.59}} & 81.58\textsuperscript{\textcolor{numpertdeltadecline}{-11.60}} & 87.71\textsuperscript{\textcolor{numpertdeltadecline}{-5.47}} \\
 & \mmasked & 94.96\textsuperscript{\textcolor{numpertdeltaimprove}{+1.40}} & 94.44\textsuperscript{\textcolor{numpertdeltaimprove}{+0.88}} & 90.53\textsuperscript{\textcolor{numpertdeltadecline}{-3.03}} & 93.21\textsuperscript{\textcolor{numpertdeltadecline}{-0.35}} \\
 & \mrandom & 58.27\textsuperscript{\textcolor{numpertdeltadecline}{-36.43}} & 94.91\textsuperscript{\textcolor{numpertdeltaimprove}{+0.21}} & 18.42\textsuperscript{\textcolor{numpertdeltadecline}{-76.28}} & 58.90\textsuperscript{\textcolor{numpertdeltadecline}{-35.80}} \\
\midrule
\Block{7-1}{\rotate \groupfourtable} & \mnooriginal & 92.81\textsuperscript{\textcolor{numpertdeltadecline}{-1.89}} & 94.44\textsuperscript{\textcolor{numpertdeltadecline}{-0.26}} & 95.79\textsuperscript{\textcolor{numpertdeltaimprove}{+1.09}} & 94.49\textsuperscript{\textcolor{numpertdeltadecline}{-0.21}} \\
 & \mnoapproximation & 53.96\textsuperscript{\textcolor{numpertdeltadecline}{-37.71}} & 89.81\textsuperscript{\textcolor{numpertdeltadecline}{-1.86}} & 93.16\textsuperscript{\textcolor{numpertdeltaimprove}{+1.49}} & 81.83\textsuperscript{\textcolor{numpertdeltadecline}{-9.84}} \\
 & \mnonumeration  & 95.68\textsuperscript{\textcolor{numpertdeltaimprove}{+2.12}} & 93.06\textsuperscript{\textcolor{numpertdeltadecline}{-0.50}} & 95.26\textsuperscript{\textcolor{numpertdeltaimprove}{+1.70}} & 94.50\textsuperscript{\textcolor{numpertdeltaimprove}{+0.94}} \\
 & \mnorange  & 96.40\textsuperscript{\textcolor{numpertdeltaimprove}{+0.19}} & \textbf{95.83}\textsuperscript{\textcolor{numpertdeltadecline}{-0.38}} & 65.79\textsuperscript{\textcolor{numpertdeltadecline}{-30.42}} & 85.50\textsuperscript{\textcolor{numpertdeltadecline}{-10.71}} \\
 & \mnonegation  & 90.65\textsuperscript{\textcolor{numpertdeltadecline}{-1.02}} & 91.20\textsuperscript{\textcolor{numpertdeltadecline}{-0.47}} & 91.58\textsuperscript{\textcolor{numpertdeltadecline}{-0.09}} & 91.19\textsuperscript{\textcolor{numpertdeltadecline}{-0.48}} \\
 & \mnomasked  & 93.53\textsuperscript{\textcolor{numpertdeltadecline}{-0.03}} & 93.06\textsuperscript{\textcolor{numpertdeltadecline}{-0.50}} & 95.79\textsuperscript{\textcolor{numpertdeltaimprove}{+2.23}} & 94.13\textsuperscript{\textcolor{numpertdeltaimprove}{+0.57}} \\
 & \mnorandom  & \textbf{97.84}\textsuperscript{\textcolor{numpertdeltaimprove}{+2.01}} & \textbf{95.83} & \textbf{96.32}\textsuperscript{\textcolor{numpertdeltaimprove}{+0.49}} & \textbf{96.51}\textsuperscript{\textcolor{numpertdeltaimprove}{+0.68}} \\

\bottomrule
\end{NiceTabular}
}
\caption{\footnotesize \ttt\ accuracy (\%) on label-preserving perturbations of originally \texttt{True} claims (Negation, Random, Masked are \texttt{False}-only and excluded). Deltas vs.\ the unperturbed original; M.\ Avg.\ is the micro-average of accuracy across perturbation types.}
\label{tab:ttt}
\end{table}


%% file: tables/results_q4_k_m/generalization.tex
\begin{table}[ht!]
\centering
\definecolor{numpertdeltaimprove}{RGB}{0,100,0}
\definecolor{numpertdeltadecline}{RGB}{180,35,24}
\small
\resizebox{\columnwidth}{!}{
\begin{tabular}{l|cccc}
\toprule
\textbf{Model} & \textbf{T1} & \textbf{T2} & \textbf{T3} & \textbf{M. Avg} \\
\midrule
\multicolumn{5}{l}{\textbf{AVeriTeC 2.0 dev}} \\
\midrule
\mbase            & \textbf{95.65} & 95.09 & 96.83 & \textbf{95.31} \\
 \mnoapproximation & 89.72\textsuperscript{\textcolor{numpertdeltadecline}{$-$5.93}} & 93.73\textsuperscript{\textcolor{numpertdeltadecline}{$-$1.37}} & \textbf{97.20}\textsuperscript{\textcolor{numpertdeltaimprove}{$+$0.37}} & 92.13\textsuperscript{\textcolor{numpertdeltadecline}{$-$3.18}} \\
\mallperturb      & 90.12\textsuperscript{\textcolor{numpertdeltadecline}{$-$5.53}} & \textbf{95.20}\textsuperscript{\textcolor{numpertdeltaimprove}{$+$0.11}} & \textbf{97.20}\textsuperscript{\textcolor{numpertdeltaimprove}{$+$0.37}} & 93.20\textsuperscript{\textcolor{numpertdeltadecline}{$-$2.11}} \\
\midrule
\multicolumn{5}{l}{\textbf{CLEF 2025 Spanish}} \\
\midrule
\mbase            & \textbf{95.51} & 97.17 & 97.40 & 96.44 \\
 \mnoapproximation & 94.83\textsuperscript{\textcolor{numpertdeltadecline}{$-$0.68}} & \textbf{99.64}\textsuperscript{\textcolor{numpertdeltaimprove}{+2.47}} & \textbf{99.54}\textsuperscript{\textcolor{numpertdeltaimprove}{+2.14}} & \textbf{97.48}\textsuperscript{\textcolor{numpertdeltaimprove}{+1.04}} \\
\mallperturb      & 94.71\textsuperscript{\textcolor{numpertdeltadecline}{$-$0.80}} & 99.62\textsuperscript{\textcolor{numpertdeltaimprove}{+2.45}} & 99.44\textsuperscript{\textcolor{numpertdeltaimprove}{+2.04}} & 97.36\textsuperscript{\textcolor{numpertdeltaimprove}{+0.92}} \\
\bottomrule
\vspace{-5pt}
\end{tabular}}
\caption{\footnotesize Cross-dataset generalisation accuracy (\%). \textbf{T1} = value-preserving; \textbf{T2} = value-changing; \textbf{T3} = random/obfuscation. Deltas vs.\ \mbase. M. Avg. is the micro-averages of accuracy of \textbf{T1}, \textbf{T2} and \textbf{T3}.}
\label{tab:generalization_results}
\end{table}

%% file: tables/results_q4_k_m/vitaminc_crossdataset.tex
\begin{table*}[ht!]
\centering
\definecolor{numpertdeltaimprove}{RGB}{0,100,0}
\definecolor{numpertdeltadecline}{RGB}{180,35,24}
\small
\begin{tabular}{l|cc|cc|ccccc}
\toprule
 & \multicolumn{2}{c|}{\textbf{VitaminC}} & \multicolumn{2}{c|}{\textbf{QuanTemp}} & \multicolumn{5}{c}{\textbf{QuanTemp by perturbation type}} \\
\textbf{Training data} & \textbf{Acc.} & \textbf{Pair} & \textbf{Acc.} & \textbf{M-F1} & \textbf{Orig.} & \textbf{T1} & \textbf{T2} & \textbf{Mask} & \textbf{Rand} \\
\midrule
None (\mbase)          & 90.96 & 84.25 & --- & --- & --- & --- & --- & --- & --- \\
QuanTemp (\mallperturb) & 78.35 & 54.45 & 97.52 & 94.62 & 96.13 & 94.95 & 98.80 & \textbf{100.00} & \textbf{99.86} \\
VitaminC (\mvitaminc)   & 98.05 & 96.56 & 80.86 & 69.80 & 79.29 & 72.69 & 92.26 & 57.88 & 95.12 \\
Both (\mmixed)          & \textbf{98.25} & \textbf{97.15} & \textbf{98.20} & \textbf{96.10} & \textbf{96.36} & \textbf{96.45} & \textbf{99.45} & \textbf{100.00} & 99.72 \\
\bottomrule
\end{tabular}
\caption{\footnotesize Each training regime on both test sets (VitaminC:
11,887 instances, 5,227 pairs; QuanTemp: 6,175). Specialising on either
corpus alone costs 17--20\,pp on the other, so neither competency is a
by-product of the other; training on both recovers it and beats each
specialist on its own test set ($+$0.68\,pp QuanTemp, $p =
1.0\mathrm{e}{-4}$; $+$0.59\,pp pair accuracy VitaminC, $p = 0.002$).
\textbf{T1} $=$ value-preserving; \textbf{T2} $=$ value-changing;
\textbf{Mask}/\textbf{Rand} are the obfuscation variants.}
\label{tab:vitaminc_crossdataset}
\end{table*}

%% file: emnlp2026-numpert-finetune-06.tex
\section{Discussion}
\label{sec:discussion}

\textit{RQ1: Does adversarial fine-tuning improve robustness to label-flipping perturbations over clean-data fine-tuning?} Yes: \mnoperturb\ is near-perfect on value-preserving evaluations yet collapses to 5--13\% across every label-flipping perturbation, an 83--90\% drop. Standard fine-tuning thus teaches the model to confirm the original claim's surface form rather than verify it against evidence: without adversarial exposure, it never sees a plausible-looking claim that is numerically inconsistent with the evidence. All QLoRA models in \grouptwo\ overcome this, matching or exceeding their \ttt\ performance on every \ttf\ perturbation.

\ftf\ results are largely unaffected: false claims are already in the majority, and there is no truth-confirming prior to override. In \ttt, all-perturbs variants lose 2.6--11.0\% on \original\ relative to \mnoperturb, likely because the adversarial objective shifts the decision boundary away from truth-confirming features; \groupfour\ largely mitigates this, as excluding original data removes the tension.

\textit{RQ2: Does training on a perturbation subset generalise, and how does training composition affect robustness?} In \ttf, every \groupthree\ model achieves ${\geq}96\%$ on its own perturbation type but collapses on at least one other. This is expected: each type exploits a distinct surface signal (digit-to-word rewriting, value-order shift, explicit negation), so training on one gives no immunity to the others. The partial exception is \mrandom, which transfers positively across \ftf: replacing any number with an arbitrary value trains a general ``distrust the number'' reflex rather than a specific pattern.

The \groupfour\ ablation shows that not all perturbation types are equally interchangeable. Excluding \masked\ or \numeration\ degrades \ttf\ performance below even the zero-shot baseline: \masked\ introduces a unique token pattern (\#\#\#) that no other type replicates, and \numeration\ requires learning digit-to-word equivalences that cannot be inferred from other rewriting patterns. By contrast, excluding \approximation\ or \random\ has a negligible effect: these signals are recoverable from cues already present in the other types.

Training composition also interacts with evaluation direction: excluding \approximation\ yields the best \ftf\ accuracy (99.23\%) but collapses in \ttt\ (53.96\%), as the model no longer recognises ``about X'' as truth-preserving. Excluding \random\ produces the strongest \ttt\ profile with little effect elsewhere. Finally, \mnooriginal, trained only on perturbed data, recovers almost all \ttt\ accuracy lost by \grouptwo, confirming that unperturbed training examples, not the perturbations themselves, shift the decision boundary.

\textit{RQ3: How do model scale and adaptation method interact with adversarial fine-tuning?} Scale matters, but fine-tuning matters more: \mallperturbsix\ beats zero-shot \mbasethirtytwo, GPT-5.4 Pro and Gemini 2.5 Flash on \ttf\ by at least 14\,pp in micro-average accuracy. Across the all-perturbs series, \ttf\ accuracy is flat at 88\% for 0.6B and 1.7B, jumps to 98\% at 4B, and saturates at 8B; the jump comes mostly from \numeration, \range, and \approximation. Both 4B and 8B outperform every \groupone\ model by at least 23\,pp.

QLoRA models in \grouptwo\ and \groupfour\ consistently outperform soft-prompt tuning. Soft prompts can shift the output distribution but do not update the weight-level representations needed to detect numerical inconsistencies; QLoRA modifies attention and feed-forward projections directly. Soft-prompt tuning still beats \mnoperturb\ on \ttf, so adversarial exposure helps, but it stays below the zero-shot baseline, suggesting the prepended vectors introduce conflicting biases rather than targeted numerical sensitivity.

\textit{RQ4: Does adversarial fine-tuning generalise across domains and languages?} Across languages, yes; on AVeriTeC, the gain is in coverage rather than accuracy (Sections~\ref{sec:generalization}--\ref{sec:significance}). On CLEF 2025 Spanish the evidence is unstructured article text, as in QuanTemp, so the shift is primarily lexical and the structural signal transfers across languages. On AVeriTeC the baseline retains a narrow edge on label-flipping, where all examples are False: its tendency to predict False inflates the score, the artefact already seen in QuanTemp \ttf.

%% file: emnlp2026-numpert-finetune-07.tex
\section{Conclusion}
\label{sec:concl}

This paper is a diagnostic study of LLM robustness to numerical claim perturbations. Standard fine-tuning teaches the model to confirm a claim's surface form rather than verify it against evidence, and accuracy on label-flipping inputs collapses to single digits. Adversarial fine-tuning on counterfactual examples recovers near-perfect accuracy, with small QLoRA-tuned models (0.6B--8B) surpassing zero-shot 32B and frontier closed-source baselines on the hardest perturbations. Composition matters more than variety: \masked\ and \numeration\ are essential, \approximation\ and \random\ redundant. Robustness also transfers across languages and domains without target-language data, suggesting a generalised numerical decision boundary rather than memorised edits.

%% file: emnlp2026-numpert-finetune-08-APPENDIX-A-revised.tex
\appendix

\section{Datasets and Perturbations}
\label{app:data}

\subsection{QuanTemp Split Statistics}
\label{app:dataset_stats}
Tables~\ref{tab:dataset_overview_train} and \ref{tab:dataset_overview_val} present the sizes of the training and validation datasets, respectively. In the training split, evidence documents contain a median of 535 words per sample; the validation split has a median of 521 words and a mean of 604; the test split (see Table~\ref{tab:dataset_overview_test}) has a median of 535 words and a mean of 648.

The maximum word count across the data splits is 6{,}943.

\input{tables/dataset-stats/dataset-stats-train}
\input{tables/dataset-stats/dataset-stats-val}

\subsection{Perturbation Type Definitions and Examples}
\label{app:perturbation_types}

Table~\ref{tab:perturbation_examples} defines each perturbation type with a concrete example. Types marked T1 generate both a value-preserving (T1) and a value-changing (T2) variant; types marked T2 or T3 generate only that variant.

\input{appendix_tables_figures/perturb_definitions}

\subsection{Cross-Dataset Evaluation Sets}
\label{app:crossdata}
\input{tables/dataset-stats/dataset-stats-generalization.tex}

\textbf{AVeriTeC 2.0 dev} \citep{Schlichtkrull:2023:NeurIPs,Akhtar:2025:FEVER} is an English claim verification dataset drawn from real-world fact-checking sources. We filter its development split\footnote{Ground-truth labels for the test split are not publicly released.} to numerical claims, retaining 119 unique claim--evidence pairs (49~\texttt{True}, 70~\texttt{False}). Evidence documents are relatively short (average 114 words), formatted as retrieved question--answer pairs rather than free-form passages. We use an LLM (Claude Opus 4.7) to draft a free-form evidence passage based on the claim, question and answer texts. We apply five perturbation categories (\numeration, \approximation, \range, \random, \negative), yielding 631 perturbed instances after removing perturbations whose text duplicates another instance (Table~\ref{tab:dataset_overview_generalization}).

\textbf{CLEF 2025 Spanish} is derived from a Spanish-language fact-checking corpus from the CLEF CheckThat! 2025 challenge~\cite{Venktesh:2025:CLEF}. After filtering to numerically grounded claims containing at least one stand-alone quantity (excluding identifier-embedded numbers such as \textit{5G} or \textit{COVID-19}), we retain 2{,}633 unique claim--evidence pairs. The split is heavily skewed towards \texttt{False} claims (200~\texttt{True} vs.\ 2{,}433~\texttt{False}), reflecting AFP-style debunking corpora that primarily publish corrections to false claims. Evidence documents are substantially longer (median 584 words, mean 642 words) as they include full fact-check articles. Applying the same five perturbation types and deduplication yields 13{,}827 perturbed instances (16{,}460 including the originals), making it the largest evaluation set in this study.

\section{Training and Inference Details}
\label{app:training}

\subsection{Training Hyperparameters}
\label{app:hyperparams}
Hyperparameters used for Hugging Face's SFTTrainer\footnote{\url{https://huggingface.co/docs/trl/sft_trainer}} configuration can be found in Table~\ref{tab:hyperparams}. Additional SFTTrainer parameters are set to their default values.

\input{appendix_tables_figures/hyperparameters}

\subsection{Soft Prompt Tuning Formulation}
\label{app:soft_prompt}

A sequence of $m$ continuous prompt vectors $P = [p_1, \ldots, p_m] \in \mathbb{R}^{m \times d}$ is prepended to the frozen input embeddings $E(x_c, x_e)$, giving the initial hidden state:
\begin{equation*}
  h^{(0)} = \bigl[P\,;\, E(x_c, x_e)\bigr]
\end{equation*}
Only $P$ and the veracity head $\{w, b\}$ are updated; $\Theta$ is fully frozen. Veracity is predicted as $\hat{y} = \sigma(w^{\top} h^{(L)}_{\text{cls}} + b)$, where $h^{(L)}_{\text{cls}}$ is the final-layer representation of the prefix position. The number of trainable parameters is $m \cdot d$, independent of model size. 

With $m=20$, soft prompts are particularly weak on label-flipping perturbations, which require rewriting (not just nudging) the decision boundary. Because this gap was already large in pilot runs, we did not pursue further soft-prompt hyperparameter tuning: closing it would require orders of magnitude more capacity, which would not address the structural limitation that soft prompts cannot modify the verification computation.

\subsection{Prompts}
\label{app:prompts}
Figure~\ref{fig:factchecking-prompt} displays the template used for the system and instruction (zero-shot) prompt used during fine-tuning and evaluation -- where \{claim\} and \{evidence\} signal the input data. In the two-shot prompt evaluation, the model is provided with an additional sentence informing it that it will receive two samples before its actual task -- all being added in the \{Two-shot sample position\}. Full two-shot examples can be accessed through our GitHub repository\footnote{\url{https://github.com/iai-group/numpert-fine-tune}}.
\input{figures/prompt}

\paragraph{\ncpfull\  (\textit{\pap})} is further examined for the \qwenbase\ base models (q4\_k\_m). The prompt includes the zero-shot instructions and additional short, trivial counterexamples illustrating why a numerical claim with each perturbation type is false.

\input{figures/NCP_example}

\subsection{Infrastructure and Hardware}
\label{app:infrastructure}

Computational resources are provided by a small GPU cluster managed with the Slurm Workload Manager. We do not report GPU-hours, as fine-tuning and inference jobs are scheduled dynamically across the available hardware rather than assigned to a fixed GPU type. The cluster is heterogeneous, comprising a small number of NVIDIA H100 80GB, NVIDIA A100 40GB, and Tesla V100 GPUs.

\section{Output Parsing}
\label{app:parsing}

\subsection{Parse-Failure Rates}
\label{app:parsing_errors}
\input{appendix_tables_figures/parse_fails_combined_table}
Table~\ref{tab:parse_failure_full_table} presents the parse-failure macro averages for the main results. Parse failures are mostly localised to non-fine-tuned models in \groupone, varying substantially across prompting settings in the main evaluation. \mbase\ produces non-trivial error rates in the \zeroshot\ setting (up to 17.90\% on T$\rightarrow$F), which \twoshot\ prompting nearly eliminates (below 3\% across directions). \mbasethirtytwo\ exhibits consistently low error rates ($\leq 4.40\%$) across all settings, suggesting that parsing failures are largely a function of model scale rather than prompting strategy. GPT-5.4 Pro and Gemini 2.5 Flash parsing errors are 2.5--4\% and 6.6--16.2\%, respectively. The unquantised counterparts exhibit higher failure rates -- \mbase\ between 40.49\% and 56.25\% and \mbasethirtytwo\ between 9.90\% and 25.78\% -- though \twoshot\ prompting improves this, reducing failures to roughly 10--15\%.

\subsection{\pap\ Under Relaxed Parsing}
\label{app:ncp}
\input{appendix_tables_figures/ncp_relaxed_table}

The high \pap\ parse-failure rate is largely a strict-parser artefact. Of the 3{,}599 \pap\ outputs rejected by the strict JSON parser, 96.8\% contain a valid \texttt{\{"label":\ true/false\}} substring embedded in a longer response (typically followed by a free-form reasoning paragraph). Under a relaxed parser that extracts the first such substring anywhere in the response, the headline \pap\ parse-failure rate drops from 58.3\% to 1.9\% (115 truly unparseable outputs out of 6{,}175), and overall \pap\ accuracy rises from 85.68\% to 90.17\%.

Table~\ref{tab:ncp_relaxed} compares the strict and relaxed \mbasepap\ rows across all three evaluation directions. The recovery is not uniform. On label-flipping (\ttf) and label-preserving False (\ftf) settings the recovered cells improve substantially (the most striking change is Negation in \ttf, $47.83 \to 72.55\%$, where strict parsing was selectively rejecting the model's correct flipping decisions). On label-preserving True (\ttt) settings the recovered cells worsen ($79.05 \to 71.50\%$ macro average), indicating that on these inputs the model's verbose-output mode produces parseable-but-incorrect predictions which strict parsing was inadvertently filtering out. We retain strict-parser numbers in the main tables as a consistent protocol across all configurations and report the relaxed numbers here for transparency.

\section{Statistical Significance}
\label{app:significance}
\input{tables/results_q4_k_m/significance}
\input{tables/results_q4_k_m/spanish_per_class}

We recompute the cross-dataset results from the stored per-instance
predictions with a paired bootstrap (10{,}000 resamples) and an exact
McNemar test (Table~\ref{tab:significance}). Because \mbase\ leaves a
substantial share of instances unparseable while the fine-tuned models
leave none, the treatment of those instances is itself a design choice,
so we report three protocols: excluding them, as in
Table~\ref{tab:generalization_results}; restricting to the instances
both models parse; and scoring them as errors.

\paragraph{CLEF Spanish.}
The gain is significant under every protocol. On the 9{,}090 instances
both models parse, \mallperturb\ improves overall accuracy by
$+$2.33\,pp (CI $[+1.99, +2.69]$, $p = 8.0\mathrm{e}{-}41$) and wins in
every perturbation direction: $+$1.96\,pp on Original
($p = 2.1\mathrm{e}{-}6$), $+$2.30\,pp on T1
($p = 1.2\mathrm{e}{-}12$), $+$2.63\,pp on T2
($p = 1.4\mathrm{e}{-}21$) and $+$2.21\,pp on T3
($p = 2.4\mathrm{e}{-}7$). Since the split is 95.8\%
\texttt{False}, accuracy sits near the majority-class rate and macro-F1
is the more informative statistic: it rises from 76.88 to 85.31
($+$8.43\,pp, CI $[+6.26, +10.72]$).
Table~\ref{tab:spanish_per_class} shows the improvement is almost
entirely in the minority \texttt{True} class, whose F1 rises by
16.4\,pp from a baseline precision of just 42.4\%; the \texttt{False}
class was already near ceiling for both models.

\paragraph{AVeriTeC.}
Here the parseable-only comparison suggests \mallperturb\ is
2.11\,pp \emph{worse} than \mbase, but that difference is not
significant once the comparison is paired over the 469 instances both
models answer ($-$1.07\,pp, CI $[-3.60, +1.46]$, $p = 0.511$). The
fine-tuned model's actual advantage on this dataset is coverage:
\mbase\ produces no parseable verdict on 281 of 750 instances
(37.5\%), against zero for \mallperturb, so under the failures-as-errors
protocol the ordering reverses to $+$33.60\,pp
($p = 4.4\mathrm{e}{-}57$). We therefore describe the AVeriTeC result as
a coverage gain with indistinguishable accuracy where both models
commit, rather than as an accuracy gain.

\section{CLEF Spanish: Per-Perturbation Breakdown}
\label{app:spanish_perturbation}
\input{tables/results_q4_k_m/spanish_perturbation}

Table~\ref{tab:generalization_results} reports CLEF Spanish only at the
level of the transition groups, which aggregate over the label
transition. On a split that is 95.8\% \texttt{False} this conceals the
effect of interest, because each group pools the instances whose label
was flipped with the far larger set whose label was already
\texttt{False}: the value-changing (T2) column combines 553 \ttf\ instances with 5{,}578
\ftf\ ones, and the T3 column 164 with 1{,}807. An aggregate therefore
tracks the \ftf\ majority, where every model is already near ceiling,
rather than the label-flipping cases the perturbations were designed to
probe.

Table~\ref{tab:spanish_perturbation} reports accuracy per perturbation
type \emph{within} each transition. Three observations follow.

\textbf{The label-flipping gain is an order of magnitude larger than the
aggregate suggests.} On \ttf, \mallperturb\ improves from 73.71\% to
97.07\% ($+$23.36\,pp) and \mnoapproximation\ to 97.49\%
($+$23.78\,pp), against the $+$2.45\,pp visible in the aggregated T2
column. Every perturbation type improves: \approximation\ by
25.16\,pp, \random\ by 20.25\,pp, \numeration\ by 16.34\,pp and
\range\ by 13.70\,pp.

\textbf{\negative\ is the baseline's worst failure and the largest
single gain.} \mbase\ reaches only 35.29\% on negated Spanish claims,
below chance for a binary decision, while both fine-tuned models reach
100.00\%. Negation inverts a percentage or rate, so the surface form
stays close to the evidence, and a model that matches on surface overlap
rather than comparing the quantity will confirm it; this mirrors the
same type being the hardest \ttf\ case for the frontier models on
QuanTemp.

\textbf{The gain is confined to the flipped cases.} On \ttt\ the
fine-tuned model is statistically indistinguishable from the baseline
(79.04\% to 79.55\%), and on \ftf\ both are near ceiling (97.44\% to
98.28\%). Adversarial fine-tuning on Spanish therefore does not make
the model better at recognising claims that remain true; it stops the
model from confirming claims whose quantity has been altered. This is
the same asymmetry visible in the per-class results of
Table~\ref{tab:spanish_per_class}, where the \texttt{True}-class gain is
driven by precision rather than recall.

\paragraph{Parse failures are concentrated in the hard cases.}
The accuracies above follow Table~\ref{tab:generalization_results} in
scoring each model on its parseable responses, which flatters \mbase\
unevenly: it fails to parse 75.6\% of \ttf\ instances, 63.3\% of \ttt\
and 42.5\% of \ftf, against zero for either fine-tuned model. Its
73.71\% on \ttf\ therefore rests on 175 of 717 instances, and on a
subset it selected by answering at all. Scoring failures as errors
instead, \mbase\ falls to 17.99\% on \ttf, 29.68\% on \ttt\ and 56.01\%
on \ftf, so the \ttf\ gap widens from $+$23.36\,pp to $+$79.08\,pp. The
true value lies between the two protocols, but under either the
label-flipping gain is far larger than the aggregated table indicates.
\section{Sensitivity to Numerical Distance}
\label{app:distance}
\input{tables/results_q4_k_m/distance_buckets}

The taxonomy of Section~\ref{sec:dataset} encodes numerical distance
only coarsely. We therefore bin each perturbed instance by the relative
distance between the original and the perturbed quantity, scoring range
perturbations as $0$ when the value falls inside the interval and
otherwise as the distance to the nearest bound. Of 4{,}579 perturbed
test instances, 3{,}794 (83\%) align to a distance;
Table~\ref{tab:distance_buckets} reports accuracy per bin.

Near-miss values are the harder case for both models, and the benefit of
adversarial fine-tuning is largest exactly there: $+$7.50\,pp within
10\% relative distance and $+$8.44\,pp between 10\% and 25\%, falling
monotonically to $+$2.91\,pp beyond 100\%. Every bin is individually
significant. This bears directly on the concern that the fine-tuned
models might be recognising perturbation artefacts rather than comparing
quantities: an artefact detector would improve uniformly across the
distance axis, or most where the edit is most conspicuous, whereas the
observed profile is the opposite and tracks the difficulty of the
numerical comparison itself.

\section{VitaminC: Evidence-Side Revisions}
\label{app:vitaminc}
\input{tables/results_q4_k_m/vitaminc_main}

\subsection{Subset Construction and Splits}
\label{app:vitaminc_details}

VitaminC contains 488,904 claim--evidence pairs, of which 163,180
(33.4\%) are annotator-written synthetic revisions rather than real
Wikipedia edits. We retain only real revisions, and only the
\texttt{SUPPORTS}/\texttt{REFUTES} labels that map onto our binary
protocol, discarding \texttt{NOT ENOUGH INFO}. We further require that
the revision is genuinely numerical, i.e.\ that the multiset of numeric
tokens differs between the two evidence snapshots, and that the claim
carries a quantitative entity (\textsc{cardinal}, \textsc{money},
\textsc{percent} or \textsc{quantity}), applying the same spaCy tagging
used to build our QuanTemp splits. Approximately 47\% of VitaminC's real revisions alter
a numeric token by this measure, consistent with the 48\% quantitative
rate reported from manual annotation \citep{Schuster:2021:NAACL}, which
is what makes the corpus a dense source of numerical contrasts.

\paragraph{Splits.} We build the subset with the filters above
applied to each official VitaminC split.
The test split yields 11,887 instances over 3,701 revisions (6,650
\texttt{True}, 5,237 \texttt{False}). Because the same claim recurs
against both snapshots, these instances include 5,227
\emph{contrastive pairs}: two evaluations of one claim whose gold labels
differ solely because a number in the evidence changed. The dev split
yields 16,434 instances over 5,143 revisions. The
splits are disjoint not only by revision and claim text but by Wikipedia
page, since VitaminC is partitioned at the page level, so no article
contributes to more than one split. We use dev only for prompt
selection, and never for model selection or reporting.

\paragraph{Evidence length.} VitaminC evidence is a single revised
sentence, with a median of 26 words (mean 28). For comparison, our QuanTemp splits have a median of 535 words
(Appendix~\ref{app:dataset_stats}) and AVeriTeC a mean of 114 words
(Appendix~\ref{app:crossdata}). The
corpus therefore isolates numerical comparison from long-context
retrieval, and it constitutes a distribution shift for models trained on
QuanTemp's much longer passages -- a shift that works \emph{against}
the fine-tuned models, making their deficit in
Table~\ref{tab:vitaminc_main} conservative rather than inflated.

\subsection{Pair Accuracy}
\label{app:pair_accuracy}

Instance accuracy understates how hard this setting is, because a model
that ignores the quantity and keys off claim--evidence surface overlap
can still score well on the supporting snapshot of every pair. We
therefore report \emph{pair accuracy}: the share of contrastive pairs for
which \emph{both} snapshots are labelled correctly. This is the direct
analogue of our label-flipping evaluation, and we additionally decompose
its errors into the two directions -- wrongly refuting a supporting
snapshot, and wrongly confirming a refuting one -- because their
asymmetry identifies the failure mode rather than merely its size.

\subsection{Training Details}
\label{app:vitaminc_training}

\paragraph{Training sample.} From the training split we draw 16,000
instances by sampling whole revisions rather than instances, so that a
contrastive pair is never divided; revisions carrying both labels are
preferred, and in the resulting sample all 4,044 revisions are
contrastive and the label distribution is balanced (7,997
\texttt{True} / 8,003 \texttt{False}). Validation is a further 2,001
instances over 508 revisions, drawn from the training split and verified
case-disjoint from it, which leaves the official dev split untouched.
The sample is sized to be comparable to the 17,258 QuanTemp training
records rather than to exhaust the 77,568 available VitaminC instances.

\paragraph{Deduplication.} Our pipeline deduplicates training samples by
claim text, which is correct for QuanTemp but destructive here: because
VitaminC pairs one claim with two evidence documents, keying on text
alone discards one snapshot of every pair, reducing the 16,000-instance
sample to 7,987 and removing precisely the contrast the corpus provides.
We therefore key deduplication on the (claim text, gold label) pair,
which retains 15,952 of 16,000 VitaminC samples (99.7\%) while changing
the QuanTemp training set by only 234 samples ($\approx$1\%). Two
samples sharing a claim but carrying opposite gold labels are
contrastive, not duplicate.

\paragraph{Optimisation.} \mvitaminc\ and \mmixed\ are trained from the
base model with the hyperparameters of Appendix~\ref{app:hyperparams}
(rank 32, batch size 5, no gradient accumulation, two epochs).
\mvitaminc\ uses a maximum sequence length of 2,048, since VitaminC's
longest prompt is 648 tokens. \mmixed\ is trained on the union of the
QuanTemp all-perturbations recipe and the VitaminC sample, 34,567
samples in total, at a maximum sequence length of 8,192: the mixed
corpus has a median prompt length of 450 tokens and a 99th percentile of
3,521, and only 0.03\% of samples exceed 8,192. For the mixed run we
leave the perturbation-ratio, T$\rightarrow$F ratio and
method-balancing constraints unset, as they are defined over QuanTemp's
perturbed-versus-original structure and would otherwise treat the
VitaminC instances as material against which to balance perturbations.

\subsection{Additional Results}
\label{app:vitaminc_results}

\paragraph{Prompting does not close the gap.}
Because the deficit looks like a decision rule rather than missing
knowledge, we ask whether prompting can recover it. Neither of our
existing strategies helps: \scot, which explicitly instructs the model
to extract, normalise and compare quantities step by step, \emph{lowers}
pair accuracy to 46.15\% and raises wrong confirmations to 44.2\%, and
\pap\ is approximately neutral at 53.74\%. We additionally designed a
\tcpfull\ (\tcp) that names the failure directly, stating that a claim
asserting a bound holds only if the evidence value satisfies that
comparison and not merely if the two numbers are close, illustrated with
invented figures so that no test instance enters the prompt. Selected on
the VitaminC dev split ($+$5.09\,pp accuracy, CI $[+4.56, +5.60]$;
$+$11.08\,pp pair accuracy, CI $[+9.99, +12.17]$), it transfers to the
test split with $+$6.81\,pp accuracy (CI $[+6.18, +7.42]$) and
$+$14.71\,pp pair accuracy (CI $[+13.45, +15.97]$), both
$p < 10^{-100}$. Even so, \tcp\ recovers only about half the deficit and
remains 15.09\,pp below \mbase\ in pair accuracy. Notably, under every
prompt the fine-tuned model emits a bare verdict of roughly 15
characters and never produces the reasoning \scot\ demands: adversarial
fine-tuning appears to have fixed the output format at the cost of
instruction-following, which is why prompts can shift its verdicts but
cannot restore deliberation.

\paragraph{Head-to-head deltas.}
The deficit of \mallperturb\ against \mbase\ is $-$12.60\,pp accuracy
(95\% CI $[-13.42, -11.77]$) and $-$29.81\,pp pair accuracy
(CI $[-31.36, -28.24]$), exact McNemar $p < 10^{-190}$.
Against \mallperturb, \mvitaminc\ gains $+$19.69\,pp accuracy and
$+$42.11\,pp pair accuracy on VitaminC. \mmixed\ is significantly better
than each specialist on that specialist's own test set: $+$0.68\,pp over
\mallperturb\ on QuanTemp (CI $[+0.34, +1.02]$,
$p = 1.0\mathrm{e}{-4}$, with 78 instances that only \mmixed\ answers
correctly against 36 for \mallperturb) and $+$0.59\,pp pair accuracy
over \mvitaminc\ on VitaminC (CI $[+0.23, +0.96]$, $p = 0.002$).

\paragraph{Validity checks.}
Three checks guard the comparison. First, we re-ran \mallperturb\ on the
QuanTemp test split through the same code path used for all VitaminC
evaluations and reproduced the published figure to within three
instances of 6,175 (97.47\% against 97.52\%), which excludes a pipeline
regression as the explanation for its VitaminC result. Second, the
VitaminC splits share no revision, no claim text and no Wikipedia page,
so the evaluation is clean by construction and the \tcp\ prompt was
selected on dev rather than test. Third, all models are evaluated at the
same 4-bit quantisation and with parse failures scored as errors, which
is conservative for \mbase: it is the only model with a non-trivial
failure rate (704 of 11,887, 5.92\%), and on parseable responses alone
it reaches 96.68\%. We note as a scope condition that \mbase\ produces
several hundred tokens of free-form reasoning per instance, whereas the
fine-tuned models emit a single verdict, so part of the baseline's
strength on this shorter-evidence corpus reflects inference-time
computation that the fine-tuned models do not spend.

\section{Unquantised Model Results}
\label{app:unquantized_results}

Tables~\ref{tab:unquant-ttf}, \ref{tab:unquant-ttt} and \ref{tab:unquant-ftf} report the same evaluation as Section~\ref{sec:results} but use unquantised 16-bit checkpoints (BF16, or FP16 on GPUs without BF16 support) rather than the q4\_k\_m quantisation used in the main paper. Models are organised into the same four groups as in the main results: \groupone\ (baselines and clean fine-tune), \grouptwo\ (all-perturbation training), \groupthree\ (single-type training), and \groupfour\ (leave-one-out training). The qualitative patterns mirror the quantised results closely across all three evaluation directions: clean fine-tuning collapses on label-flipping perturbations, all-perturbs training restores near-perfect accuracy, single-type training fails to generalise across perturbation types, and the same perturbation types remain essential (\masked, \numeration) or redundant (\approximation, \random). Per-cell numbers differ by at most a few percentage points in either direction, but the relative ordering of model groups and the qualitative conclusions of the main paper hold.

\subsection{\texttt{True} $\rightarrow$ \texttt{False}}
\label{sec:unquant-ttf}
\input{appendix_tables_figures/appendix_results_unquant/T-_F}

\groupone\ collapses on \negative\ and \masked, the absolute outlier being \mnoperturb\ with 94.70\% on originals but only 9.42\% micro-average ($-$85.28\,pp) and 79--87\% drops across every perturbed category -- closely tracking the collapse seen in the quantised setting (Table~\ref{tab:ttf}). \grouptwo\ is uniformly more robust, with \mallperturb\ showing the lowest accuracy fluctuations in its group. \groupthree\ reaches perfect accuracy on its fine-tuned perturbation type but collapses elsewhere, again matching the single-type generalisation pattern reported in the main results. \groupfour\ achieves the best overall accuracies: \mnomasked\ drops $-$75.85\% on its held-out perturbation type (close to the $-$80.13\% reported for the quantised counterpart), while \mnoapproximation\ and \mnorange\ \emph{improve} on their held-out type, confirming that the \approximation\ and \range\ signals are recoverable from compositional cues in the remaining training mix.

\subsection{\texttt{True} $\rightarrow$ \texttt{True}}
\label{sec:unquant-ttt}
\input{appendix_tables_figures/appendix_results_unquant/T-_T}

Table~\ref{tab:unquant-ttt} presents the results for the \ttt\ evaluation. In sharp contrast to \ttf, \groupone\ dominates: \mbase\ reaches a micro-average of 96.69\%, while \grouptwo\ trails in the 83--91\% range. This is the same direction observed in the quantised results (Table~\ref{tab:ttt}), where all-perturbs training loses ground on label-preserving inputs because the adversarial objective shifts the decision boundary away from truth-confirming features. \groupthree\ and \groupfour\ instabilities concentrate on the \range\ perturbation type, again consistent with the quantised findings.

\subsection{\texttt{False} $\rightarrow$ \texttt{False}}
\label{sec:unquant-ftf}
\input{appendix_tables_figures/appendix_results_unquant/F-_F}

In the \ftf\ evaluation (Table~\ref{tab:unquant-ftf}), most models achieve robust accuracies of 96--99\%, indicating that fine-tuning largely resolves the small blind spots observed in \groupone -- identical to the pattern in the quantised setting (Table~\ref{tab:ftf}). The leave-one-out models show little degradation on their held-out perturbation types, with \mnoapproximation\ and \mnorange\ both scoring near saturation.

\paragraph{Summary.} Across all three evaluation directions, the unquantised results reproduce the qualitative findings of the q4\_k\_m experiments: the same model groups, perturbation types, and headline conclusion, that adversarial fine-tuning mitigates label-flipping brittleness at a small cost to label-preserving accuracy, hold independently of quantisation.

%% file: tables/dataset-stats/dataset-stats-train.tex
\begin{table}[ht!]
\small
\centering
\begin{tabular}{lcccc}

\toprule
\textbf{Category} & \textbf{T$\rightarrow$T} & \textbf{T$\rightarrow$F} & \textbf{F$\rightarrow$F} & \textbf{Total} \\
\midrule
\textbf{\numeration} & 775 & 775 & 3872 & 5422 \\
\textbf{\approximation} & 539 & 539 & 3010 & 4088 \\
\textbf{\range} & 931 & 931 & 4868 & 6730 \\
\textbf{\masked} & \xmark & 775 & 1939 & 2714 \\
\textbf{\random} & \xmark & 775 & 1939 & 2714 \\
\textbf{\negative} & \xmark & 218 & 294 & 512 \\
\bottomrule
\end{tabular}
\caption{\footnotesize QuanTemp \textbf{train} split: sample counts by perturbation type and transition direction. Original claims: 931 \texttt{True}, 2{,}434 \texttt{False}.}
\label{tab:dataset_overview_train}
\end{table}

%% file: tables/dataset-stats/dataset-stats-val.tex
\begin{table}[ht!]
\small
\centering
\begin{tabular}{lcccc}

\toprule
\textbf{Category} & \textbf{T$\rightarrow$T} & \textbf{T$\rightarrow$F} & \textbf{F$\rightarrow$F} & \textbf{Total} \\
\midrule
\textbf{\numeration} & 273 & 273 & 1206 & 1752 \\
\textbf{\approximation} & 197 & 197 & 938 & 1332 \\
\textbf{\range} & 324 & 324 & 1494 & 2142 \\
\textbf{\masked} & \xmark & 273 & 603 & 876 \\
\textbf{\random} & \xmark & 273 & 603 & 876 \\
\textbf{\negative} & \xmark & 75 & 95 & 170 \\
\bottomrule
\end{tabular}
\caption{\footnotesize QuanTemp \textbf{validation} split: sample counts by perturbation type and transition direction. Original claims: 324 \texttt{True}, 747 \texttt{False}.}
\label{tab:dataset_overview_val}
\end{table}

%% file: appendix_tables_figures/perturb_definitions.tex
\begin{table*}[ht]
\centering\small
\begin{tabular}{llp{4.2cm}p{4.2cm}c}
\toprule
\textbf{Type} & \textbf{Label} & \textbf{Original claim fragment} & \textbf{Perturbed fragment} & \textbf{Variant} \\
\midrule
Numeration     & T1 & ``fell to \textbf{3.5\%}'' & ``fell to \textbf{three point five percent}'' & T1 (preserving) \\
               &    &                              & ``fell to \textbf{thirty-five percent}''       & T2 (changing) \\
\addlinespace
Approximation  & T1 & ``matched \textbf{50 years}'' & ``matched \textbf{about 50 years}'' & T1 (preserving) \\
               &    &                                & ``matched \textbf{about 5 years}''  & T2 (changing) \\
\addlinespace
Range          & T1 & ``over \textbf{30 million} filed'' & ``over \textbf{25--35 million} filed'' & T1 (preserving) \\
               &    &                                    & ``over \textbf{10--15 million} filed'' & T2 (changing) \\
\addlinespace
Masked        & T3 & ``rate of \textbf{3.5\%}'' & ``rate of \textbf{\#\#\#\#}'' & T3 only \\
\addlinespace
Random         & T3 & ``fell by \textbf{12 points}'' & ``fell by \textbf{47 points}'' & T3 only \\
\addlinespace
Negation       & T2 & ``quarterly earnings was \textbf{5\%}'' & ``quarterly earnings was \textbf{-5\%}'' & T2 only \\
\bottomrule
\end{tabular}
\caption{\footnotesize Perturbation type definitions and examples~\citep{Aarnes:2025:IJCNLP}. T1 = value-preserving; T2 = value-changing; T3 = random replacement and masked obfuscation.}
\label{tab:perturbation_examples}
\end{table*}

%% file: tables/dataset-stats/dataset-stats-generalization.tex
\begin{table}[ht!]
\small

\centering
\resizebox{\columnwidth}{!}{%
\begin{tabular}{lcccc}
\toprule
 & \multicolumn{2}{c}{\textbf{AVeriTeC 2.0 dev}} & \multicolumn{2}{c}{\textbf{CLEF 2025 Spanish}} \\
\cmidrule(lr){2-3}\cmidrule(lr){4-5}
\textbf{Category} & \textbf{True} & \textbf{False} & \textbf{True} & \textbf{False} \\
\midrule
\textbf{Num.}    &  74 &  124 &   328 & 3{,}614 \\
\textbf{Approx.} &  64 &   90 &   322 & 3{,}470 \\
\textbf{Range}   &  64 &   94 &   317 & 3{,}397 \\
\textbf{Rand.}   &  45 &   62 &   164 & 1{,}807 \\
\textbf{Neg.}    &   5 &    9 &    70 &   338   \\
\midrule
\textbf{Total}   & 252 &  379 & 1{,}201 & 12{,}626 \\
\bottomrule
\end{tabular}
}
\caption{\footnotesize Evaluated perturbed instances for the cross-dataset generalisation splits, by label of the source claim. Numeration, Approximation and Range each contribute a value-preserving (T1) and a value-changing (T2) variant per claim, both of which remain \texttt{False} for \texttt{False} source claims; only the \random{} obfuscation variant is evaluated, and perturbations whose text duplicates another instance are removed. \textbf{AVeriTeC 2.0 dev}: 119 claims (49T / 70F), 750 instances including originals; \textbf{CLEF 2025 Spanish}: 2{,}633 claims (200T / 2{,}433F), 16{,}460 instances including originals.}
\label{tab:dataset_overview_generalization}
\end{table}

%% file: appendix_tables_figures/hyperparameters.tex
\begin{table}[h]
\centering\small
\resizebox{\columnwidth}{!}{%
\begin{tabular}{ll}
\toprule
\textbf{Hyperparameter} & \textbf{Value} \\
\midrule
    Epochs & 2 \\
    Batch size (per device) & 5 \\
    Gradient accumulation & 1 \\
    Learning rate & $1\times10^{-4}$ \\
    LR scheduler & Cosine \\
    Warmup steps & 50 \\
    Optimiser & AdamW (8-bit) \\
    Weight decay & 0.01 \\
    Max sequence length & 16{,}384 \\
    Precision & bfloat16 / fp16 \\
    Quantisation & 4-bit NF4 \\
    LoRA rank $r$ & 32 \\
    LoRA $\alpha$ & 16 \\
    LoRA dropout & 0 \\
    LoRA target modules & q, k, v, o, gate, up, down proj \\
\bottomrule
\end{tabular}
}
\caption{\footnotesize QLoRA training hyperparameters.}
\label{tab:hyperparams}
\end{table}

%% file: figures/prompt.tex
\begin{figure}[t]
\centering
\begin{tcolorbox}[
    width=\columnwidth,
    colback=gray!3,
    colframe=gray!45,
    boxrule=0.5pt,
    arc=2mm,
    left=1.5mm,
    right=1.5mm,
    top=1.5mm,
    bottom=1.5mm,
    fontupper=\small
]
\textbf{System prompt}\\[-1mm]
\begin{quote}
You are a professional fact-checker, your task is to classify whether the given claim is true or false based on the evidence text provided.
\end{quote}

\vspace{1mm}
\textbf{Instruction prompt}\\[-1mm]
\begin{quote}
Given the claim and evidence provided, classify the claim as \texttt{\{\{"label": true\}\}} if it is true, and \texttt{\{\{"label": false\}\}} if it is false.

\vspace{1mm}
\{Two-shot or \pap\ examples\}

\vspace{1mm}
\textbf{Claim:} \texttt{\{claim\}}

\textbf{Evidence:} \texttt{\{evidence\}}
\end{quote}
\end{tcolorbox}
\caption{\footnotesize Prompt template for binary fact-checking. The model receives a claim and an evidence passage, then returns a JSON label indicating whether the claim is true or false.}
\label{fig:factchecking-prompt}
\end{figure}

%% file: figures/NCP_example.tex
\begin{figure}[t]
\centering
\begin{tcolorbox}[
    width=\columnwidth,
    colback=gray!3,
    colframe=gray!45,
    boxrule=0.5pt,
    arc=2mm,
    left=1.5mm,
    right=1.5mm,
    top=1.5mm,
    bottom=1.5mm,
    fontupper=\small
]
\textbf{Snippet of NCP Prompt}\\[-1mm]
\begin{quote}
The numbers in the evidence may not match the claim. For example:

\vspace{1mm}
Claim: The Eiffel Tower is three hundred and fifty-one meters tall.

\vspace{1mm}
Evidence: The Eiffel Tower is 330 meters tall.

\vspace{1mm}
Reasoning: The claim states that the Eiffel Tower is three hundred and fifty-one meters tall (351 m), but the evidence indicates it is 330 meters tall. Since the numbers do not match, the claim is false.

\vspace{1mm}
\{"label": false\}
\end{quote}
\end{tcolorbox}
\caption{\footnotesize Approximation NCP example. The full prompt provides one example per perturbation type and is added to the instruction prompt.}
\label{fig:ncp-prompt}
\end{figure}

%% file: appendix_tables_figures/parse_fails_combined_table.tex
\begin{table}[t]
  \centering
  \resizebox{\columnwidth}{!}{%
  \begin{tabular}{lrrr}
    \toprule
    \textbf{Model} & T$\to$F & T$\to$T & F$\to$F \\
    \midrule
    Gemini           & 	14.38 & 16.19 &  6.64 \\
    GPT-5.4          &  3.00 &  4.00 &  2.50 \\
    \midrule
    \multicolumn{4}{l}{\textbf{Quantised (Q4\_K\_M)}} \\
    \mbase           & 17.90 & 10.30 & 11.90 \\
    \mbasetwoshot    &  2.40 &  2.80 &  0.50 \\
    \mbasethirtytwo  &  4.40 &  3.70 &  1.30 \\
    \mptfulldata      &  9.70 &  0.00 &  0.00 \\
    \midrule
    \multicolumn{4}{l}{\textbf{Unquantised}} \\
    \mbase           & 56.25 & 40.49 & 55.54 \\
    \mbasetwoshot    & 14.93 & 10.14 & 12.39 \\
    \mbasethirtytwo  & 25.78 & 16.16 &  9.90 \\
    \mmasked         &  6.62 &  4.00 &  4.34 \\
    \mrandom         &  --   &  8.25 & 10.20 \\
    \bottomrule
  \end{tabular}
  }
  \caption{\footnotesize Parse failure rates (\%) by split and quantisation.}
  \label{tab:parse_failure_full_table}
\end{table}

%% file: appendix_tables_figures/ncp_relaxed_table.tex
\begin{table}[ht]
\centering\small
\begin{tabular}{lrrr}
\toprule
\textbf{Column} & \textbf{Strict} & \textbf{Relaxed} & $\Delta$ \\
\midrule
\multicolumn{4}{l}{\textbf{\ttf}} \\
\midrule
Orig.    & 76.16 & 73.93 & $-$2.23 \\
Approx.  & 89.66 & 94.93 & $+$5.27 \\
Num.     & 81.00 & 85.92 & $+$4.92 \\
Range    & 87.23 & 91.94 & $+$4.71 \\
Neg.     & 47.83 & 72.55 & $+$24.72 \\
Mask.    & 30.53 & 33.15 & $+$2.62 \\
Rand.    & 86.46 & 92.49 & $+$6.03 \\
M.\,Avg. & 72.20 & 78.50 & $+$6.30 \\
\midrule
\multicolumn{4}{l}{\textbf{\ftf}} \\
\midrule
Orig.    & 91.74 & 94.79 & $+$3.05 \\
Approx.  & 94.53 & 95.90 & $+$1.37 \\
Num.     & 93.68 & 95.98 & $+$2.30 \\
Range    & 77.27 & 87.32 & $+$10.05 \\
Neg.     & 94.74 & 96.70 & $+$1.96 \\
Mask.    & 89.95 & 92.37 & $+$2.42 \\
Rand.    & 98.82 & 99.00 & $+$0.18 \\
M.\,Avg. & 91.28 & 94.55 & $+$3.27 \\
\midrule
\multicolumn{4}{l}{\textbf{\ttt}} \\
\midrule
Approx.  & 64.63 & 59.56 & $-$5.07 \\
Num.     & 83.94 & 79.81 & $-$4.13 \\
Range    & 84.03 & 75.14 & $-$8.89 \\
M.\,Avg. & 79.05 & 71.50 & $-$7.55 \\
\bottomrule
\end{tabular}
\caption{\footnotesize \mbasepap\ accuracy (\%) under strict (used in main tables) vs.\ relaxed JSON parsing. Relaxed parsing extracts the first \texttt{\{"label":\ldots\}} substring anywhere in the response, recovering 96.8\% of strict-parser rejections (overall parse-failure rate drops from 58.3\% to 1.9\%). Recovery improves \ttf\ and \ftf\ cells but worsens \ttt\ cells.}
\label{tab:ncp_relaxed}
\end{table}

%% file: tables/results_q4_k_m/significance.tex
\begin{table*}[ht!]
\centering
\small
\resizebox{\textwidth}{!}{%
\begin{tabular}{ll|rrr|rr}
\toprule
\textbf{Dataset} & \textbf{Protocol} & \textbf{\mbase} & \textbf{\mallperturb} & \textbf{$\Delta$ (pp)} & \textbf{95\% CI (pp)} & \textbf{McNemar $p$} \\
\midrule
\multirow{3}{*}{CLEF Spanish} & Parseable-only (reported) & 96.44 & 97.36 & $+$0.93 & $[+0.53, +1.35]$ & --- \\
 & Paired ($n = 9{,}090$) & 96.44 & 98.77 & $+$2.33 & $[+1.99, +2.69]$ & $8.0\mathrm{e}{-}41$ \\
 & Failures as errors & 53.26 & 97.36 & $+$44.11 & $[+43.34, +44.86]$ & $< 10^{-300}$ \\
\midrule
\multirow{3}{*}{AVeriTeC} & Parseable-only (reported) & 95.31 & 93.20 & $-$2.11 & $[-4.51, +0.33]$ & --- \\
 & Paired ($n = 469$) & 95.31 & 94.24 & $-$1.07 & $[-3.60, +1.46]$ & $0.511$ \\
 & Failures as errors & 59.60 & 93.20 & $+$33.60 & $[+29.87, +37.33]$ & $4.4\mathrm{e}{-}57$ \\
\bottomrule
\end{tabular}%
}
\caption{\footnotesize Significance of the cross-dataset gains
(overall accuracy, \%). Deltas are \mallperturb\ minus \mbase, with a
10{,}000-resample paired bootstrap and an exact McNemar test on stored
per-instance predictions. The three protocols differ only in how
\mbase's unparseable responses are handled: excluded (as in
Table~\ref{tab:generalization_results}), restricted to the instances
\emph{both} models parse, or scored as errors. On CLEF Spanish the gain
is significant under every protocol; on AVeriTeC the apparent
parseable-only deficit is not significant once the comparison is paired
($p = 0.511$), and the fine-tuned model's real advantage there is
coverage: \mbase\ returns no parseable verdict on 281 of 750 instances
(37.5\%) against zero for \mallperturb.}
\label{tab:significance}
\end{table*}

%% file: tables/results_q4_k_m/spanish_per_class.tex
\begin{table}[ht!]
\centering
\small
\resizebox{\columnwidth}{!}{
\begin{tabular}{l|ccc|ccc|c}
\toprule
 & \multicolumn{3}{c|}{\texttt{True}} & \multicolumn{3}{c|}{\texttt{False}} & \\
\textbf{Model} & \textbf{P} & \textbf{R} & \textbf{F1} & \textbf{P} & \textbf{R} & \textbf{F1} & \textbf{M-F1} \\
\midrule
\mbase            & 42.4 & 80.9 & 55.6 & \textbf{99.4} & 96.9 & 98.1 & 76.88 \\
\mnoapproximation & \textbf{65.7} & \textbf{82.5} & \textbf{73.1} & 99.2 & \textbf{98.1} & \textbf{98.7} & \textbf{85.89} \\
\mallperturb      & 64.4 & 81.6 & 72.0 & 99.2 & 98.0 & 98.6 & 85.31 \\
\bottomrule
\end{tabular}}
\caption{\footnotesize Per-class results on CLEF 2025 Spanish (\%).
The split is 95.8\% \texttt{False}, so overall accuracy is close to the
majority-class rate and macro-F1 (\textbf{M-F1}) is the informative
figure: fine-tuning improves it by 8.4--9.0\,pp, and the improvement is
concentrated almost entirely in the minority \texttt{True} class
($+$16.4\,pp F1 for \mallperturb), where \mbase's precision is only
42.4\%. Support differs across rows because \mbase\ is scored on the
9{,}090 instances it parses (251 \texttt{True}) whereas the fine-tuned
models are scored on all 16{,}460 (684 \texttt{True}); \mbase\ fails to
parse 7{,}370 instances (44.8\%) against zero for either fine-tuned
model.}
\label{tab:spanish_per_class}
\end{table}

%% file: tables/results_q4_k_m/spanish_perturbation.tex
\begin{table*}[ht!]
\centering
\definecolor{numpertdeltaimprove}{RGB}{0,100,0}
\definecolor{numpertdeltadecline}{RGB}{180,35,24}
\small
\setlength{\tabcolsep}{4pt}
\resizebox{\textwidth}{!}{%
\begin{tabular}{l|ccccc|c}
\toprule
\textbf{Model} & \textbf{Approx.} & \textbf{Num.} & \textbf{Range} & \textbf{Neg.} & \textbf{Rand.} & \textbf{M. Avg.} \\
\midrule
\multicolumn{7}{l}{\textit{\ttf\ (label-flipping)}} \\
\midrule
\mbase & 72.97 & 80.00 & 82.50 & 35.29 & 76.09 & 73.71 \\
\mnoapproximation & 98.76\textsuperscript{\textcolor{numpertdeltaimprove}{$+$25.78}} & 97.56\textsuperscript{\textcolor{numpertdeltaimprove}{$+$17.56}} & 95.57\textsuperscript{\textcolor{numpertdeltaimprove}{$+$13.07}} & 100.00\textsuperscript{\textcolor{numpertdeltaimprove}{$+$64.71}} & 96.95\textsuperscript{\textcolor{numpertdeltaimprove}{$+$20.86}} & 97.49\textsuperscript{\textcolor{numpertdeltaimprove}{$+$23.78}} \\
\mallperturb & 98.14\textsuperscript{\textcolor{numpertdeltaimprove}{$+$25.16}} & 96.34\textsuperscript{\textcolor{numpertdeltaimprove}{$+$16.34}} & 96.20\textsuperscript{\textcolor{numpertdeltaimprove}{$+$13.70}} & 100.00\textsuperscript{\textcolor{numpertdeltaimprove}{$+$64.71}} & 96.34\textsuperscript{\textcolor{numpertdeltaimprove}{$+$20.25}} & 97.07\textsuperscript{\textcolor{numpertdeltaimprove}{$+$23.36}} \\
\textit{(n)} & \textit{161} & \textit{164} & \textit{158} & \textit{70} & \textit{164} & \textit{717} \\
\midrule
\multicolumn{7}{l}{\textit{\ttt\ (label-preserving)}} \\
\midrule
\mbase & 78.43 & 82.26 & 75.93 & -- & -- & 79.04 \\
\mnoapproximation & 64.60\textsuperscript{\textcolor{numpertdeltadecline}{$-$13.84}} & 85.98\textsuperscript{\textcolor{numpertdeltaimprove}{$+$3.72}} & 89.31\textsuperscript{\textcolor{numpertdeltaimprove}{$+$13.38}} & -- & -- & 79.96\textsuperscript{\textcolor{numpertdeltaimprove}{$+$0.92}} \\
\mallperturb & 77.64\textsuperscript{\textcolor{numpertdeltadecline}{$-$0.79}} & 79.27\textsuperscript{\textcolor{numpertdeltadecline}{$-$2.99}} & 81.76\textsuperscript{\textcolor{numpertdeltaimprove}{$+$5.84}} & -- & -- & 79.55\textsuperscript{\textcolor{numpertdeltaimprove}{$+$0.50}} \\
\textit{(n)} & \textit{161} & \textit{164} & \textit{159} & \textit{--} & \textit{--} & \textit{484} \\
\midrule
\multicolumn{7}{l}{\textit{\ftf\ (already false)}} \\
\midrule
\mbase & 97.85 & 97.55 & 96.31 & 97.28 & 98.39 & 97.44 \\
\mnoapproximation & 99.14\textsuperscript{\textcolor{numpertdeltaimprove}{$+$1.29}} & 98.34\textsuperscript{\textcolor{numpertdeltaimprove}{$+$0.79}} & 96.53\textsuperscript{\textcolor{numpertdeltaimprove}{$+$0.22}} & 100.00\textsuperscript{\textcolor{numpertdeltaimprove}{$+$2.72}} & 99.78\textsuperscript{\textcolor{numpertdeltaimprove}{$+$1.39}} & 98.32\textsuperscript{\textcolor{numpertdeltaimprove}{$+$0.88}} \\
\mallperturb & 98.33\textsuperscript{\textcolor{numpertdeltaimprove}{$+$0.48}} & 98.51\textsuperscript{\textcolor{numpertdeltaimprove}{$+$0.96}} & 97.06\textsuperscript{\textcolor{numpertdeltaimprove}{$+$0.75}} & 100.00\textsuperscript{\textcolor{numpertdeltaimprove}{$+$2.72}} & 99.72\textsuperscript{\textcolor{numpertdeltaimprove}{$+$1.33}} & 98.28\textsuperscript{\textcolor{numpertdeltaimprove}{$+$0.84}} \\
\textit{(n)} & \textit{3470} & \textit{3614} & \textit{3397} & \textit{338} & \textit{1807} & \textit{12626} \\
\bottomrule
\end{tabular}%
}
\caption{\footnotesize CLEF 2025 Spanish accuracy (\%) by perturbation type \emph{within} each label transition. Deltas are against \mbase\ in the same column. Accuracy is computed on each model's parseable responses, as in Table~\ref{tab:generalization_results}. \textbf{Neg.} and \textbf{Rand.} produce \texttt{False} labels by construction and so cannot appear under \ttt. Reporting within transitions separates the 717 label-flipping instances from the 15{,}059 \ftf\ ones that dominate this split; the aggregated columns of Table~\ref{tab:generalization_results} pool them and understate the label-flipping gain by an order of magnitude.}
\label{tab:spanish_perturbation}
\end{table*}

%% file: tables/results_q4_k_m/distance_buckets.tex
\begin{table}[ht!]
\centering
\small
\resizebox{\columnwidth}{!}{%
\begin{tabular}{l|r|rrr}
\toprule
\textbf{Relative distance} & \textbf{$n$} & \textbf{\mbase} & \textbf{\mallperturb} & \textbf{$\Delta$} \\
\midrule
0 (value preserved) & 1{,}373 & 90.17 & 96.07 & $+$5.90 \\
$(0, 10\%]$         &   399 & 87.99 & 95.49 & $+$7.50 \\
$(10\%, 25\%]$      &   345 & 90.11 & 98.55 & $+$8.44 \\
$(25\%, 50\%]$      &   661 & 94.14 & 99.39 & $+$5.26 \\
$(50\%, 100\%]$     &   513 & 94.06 & 98.83 & $+$4.77 \\
$> 100\%$           &   503 & 96.69 & 99.60 & $+$2.91 \\
\bottomrule
\end{tabular}%
}
\caption{\footnotesize Accuracy (\%) by the relative distance between
the original and the perturbed quantity, on the QuanTemp test split
(3{,}794 of 4{,}579 perturbed instances align to a distance, 83\%
coverage). \mbase\ is scored on parseable responses only. Every bin is
individually significant (exact McNemar, $p < 10^{-2}$ throughout and
$p < 10^{-7}$ in three bins). Near-miss values are the harder case for
both models, and the gain from adversarial fine-tuning is largest
exactly there ($+$7.5 to $+$8.4\,pp below 25\% distance) and smallest
for implausible replacements beyond 100\%, which is the opposite of what
a model that had merely learned to recognise perturbation artefacts
would show.}
\label{tab:distance_buckets}
\end{table}

%% file: tables/results_q4_k_m/vitaminc_main.tex
\begin{table*}[ht!]
\centering
\definecolor{numpertdeltaimprove}{RGB}{0,100,0}
\definecolor{numpertdeltadecline}{RGB}{180,35,24}
\small
\begin{tabular}{ll|cc|c|cc|c}
\toprule
 & & \multicolumn{2}{c|}{\textbf{Per instance}} & \textbf{Per pair} & \multicolumn{2}{c|}{\textbf{Pair error mode}} & \\
\textbf{Model} & \textbf{Prompt} & \textbf{Acc.} & \textbf{Macro-F1} & \textbf{Pair Acc.} & \textbf{Refutes} & \textbf{Confirms} & \textbf{PF} \\
\midrule
\multicolumn{8}{l}{\textit{No target-domain training}} \\
\midrule
\mbase & Zero-shot & 90.96 & 90.86 & 84.25 & 8.8 & 8.4 & 704 \\
\midrule
\multicolumn{8}{l}{\textit{Trained on QuanTemp claim-side perturbations}} \\
\midrule
\mallperturb & Zero-shot & 78.35\textsuperscript{\textcolor{numpertdeltadecline}{$-$12.60}} & 77.63 & 54.45\textsuperscript{\textcolor{numpertdeltadecline}{$-$29.81}} & 15.0 & 31.5 & 0 \\
\mallperturb & \scot & 74.98\textsuperscript{\textcolor{numpertdeltadecline}{$-$15.98}} & 73.20 & 46.15\textsuperscript{\textcolor{numpertdeltadecline}{$-$38.10}} & 10.7 & 44.2 & 0 \\
\mallperturb & \pap & 76.18\textsuperscript{\textcolor{numpertdeltadecline}{$-$14.78}} & 76.00 & 53.74\textsuperscript{\textcolor{numpertdeltadecline}{$-$30.51}} & 24.2 & 23.2 & 0 \\
\mallperturb & \tcp & 85.16\textsuperscript{\textcolor{numpertdeltadecline}{$-$5.80}} & 84.85 & 69.16\textsuperscript{\textcolor{numpertdeltadecline}{$-$15.09}} & 12.1 & 19.5 & 0 \\
\midrule
\multicolumn{8}{l}{\textit{Trained with VitaminC evidence-side revisions}} \\
\midrule
\mvitaminc & Zero-shot & 98.05\textsuperscript{\textcolor{numpertdeltaimprove}{$+$7.09}} & 98.02 & 96.56\textsuperscript{\textcolor{numpertdeltaimprove}{$+$12.30}} & 1.2 & 2.5 & 0 \\
\mmixed & Zero-shot & \textbf{98.25}\textsuperscript{\textcolor{numpertdeltaimprove}{$+$7.29}} & \textbf{98.23} & \textbf{97.15}\textsuperscript{\textcolor{numpertdeltaimprove}{$+$12.89}} & 1.4 & 1.7 & 0 \\
\bottomrule
\end{tabular}
\caption{\footnotesize Results on the VitaminC numerical test subset
(11,887 instances; 5,227 contrastive pairs). \textbf{Pair Acc.} is the
share of contrastive pairs for which both snapshots of a revision are
labelled correctly. \textbf{Refutes} is the share of pairs whose
supporting snapshot is wrongly refuted, and \textbf{Confirms} the share
whose refuting snapshot is wrongly confirmed; the asymmetry between them
identifies the failure mode. \textbf{PF} counts parse failures, which are
scored as errors throughout so that all rows share the same denominator.
Deltas are against \mbase. \scot~$=$~\scotfull, \pap~$=$~\ncpfull,
\tcp~$=$~\tcpfull.}
\label{tab:vitaminc_main}
\end{table*}

%% file: appendix_tables_figures/appendix_results_unquant/T-_F.tex
\begin{table*}[ht!!!]
\centering
\definecolor{numpertdeltaimprove}{RGB}{0,100,0}
\definecolor{numpertdeltadecline}{RGB}{180,35,24}
\small
\setlength{\tabcolsep}{3pt}
\resizebox{\textwidth}{!}{%
\begin{NiceTabular}{l|cccccccc}[first-col]
\toprule
 & \textbf{Model} & \textbf{Orig.} & \textbf{Approx.} & \textbf{Num.} & \textbf{Range} & \textbf{Neg.} & \textbf{Mask.} & \textbf{Rand.} & \textbf{Micro Avg.} \\
\midrule
\Block{5-1}{\rotate \grouponetable} & \mbase & 94.35 & 81.82\textsuperscript{\textcolor{numpertdeltadecline}{-12.53}} & 35.94\textsuperscript{\textcolor{numpertdeltadecline}{-58.41}} & 78.18\textsuperscript{\textcolor{numpertdeltadecline}{-16.17}} & 3.70\textsuperscript{\textcolor{numpertdeltadecline}{-90.65}} & 2.50\textsuperscript{\textcolor{numpertdeltadecline}{-91.85}} & 71.43\textsuperscript{\textcolor{numpertdeltadecline}{-22.92}} & 48.74\textsuperscript{\textcolor{numpertdeltadecline}{-45.61}} \\
 & \mbasetwoshot & 86.92 & 86.61\textsuperscript{\textcolor{numpertdeltadecline}{-0.31}} & 62.50\textsuperscript{\textcolor{numpertdeltadecline}{-24.42}} & 80.00\textsuperscript{\textcolor{numpertdeltadecline}{-6.92}} & 21.74\textsuperscript{\textcolor{numpertdeltadecline}{-65.18}} & 8.63\textsuperscript{\textcolor{numpertdeltadecline}{-78.29}} & 84.70\textsuperscript{\textcolor{numpertdeltadecline}{-2.22}} & 60.28\textsuperscript{\textcolor{numpertdeltadecline}{-26.64}} \\
 & \mbasethirtytwo & 88.24 & 85.86\textsuperscript{\textcolor{numpertdeltadecline}{-2.38}} & 66.42\textsuperscript{\textcolor{numpertdeltadecline}{-21.82}} & 85.34\textsuperscript{\textcolor{numpertdeltadecline}{-2.90}} & 29.27\textsuperscript{\textcolor{numpertdeltadecline}{-58.97}} & 7.29\textsuperscript{\textcolor{numpertdeltadecline}{-80.95}} & 83.80\textsuperscript{\textcolor{numpertdeltadecline}{-4.44}} & 61.90\textsuperscript{\textcolor{numpertdeltadecline}{-26.34}} \\
 & \mnoperturb & \textbf{94.70} & 8.63\textsuperscript{\textcolor{numpertdeltadecline}{-86.07}} & 6.94\textsuperscript{\textcolor{numpertdeltadecline}{-87.76}} & 9.47\textsuperscript{\textcolor{numpertdeltadecline}{-85.23}} & 7.69\textsuperscript{\textcolor{numpertdeltadecline}{-87.01}} & 6.94\textsuperscript{\textcolor{numpertdeltadecline}{-87.76}} & 15.28\textsuperscript{\textcolor{numpertdeltadecline}{-79.42}} & 9.42\textsuperscript{\textcolor{numpertdeltadecline}{-85.28}} \\
\midrule
\Block{4-1}{\rotate \grouptwotable} & \mallperturbsix & 81.06 & 93.53\textsuperscript{\textcolor{numpertdeltaimprove}{+12.47}} & 75.00\textsuperscript{\textcolor{numpertdeltadecline}{-6.06}} & 83.68\textsuperscript{\textcolor{numpertdeltaimprove}{+2.62}} & \textbf{100.00}\textsuperscript{\textcolor{numpertdeltaimprove}{+18.94}} & 99.54\textsuperscript{\textcolor{numpertdeltaimprove}{+18.48}} & 96.76\textsuperscript{\textcolor{numpertdeltaimprove}{+15.70}} & 90.09\textsuperscript{\textcolor{numpertdeltaimprove}{+9.03}} \\
 & \mallperturbseventeen & 85.23 & 89.93\textsuperscript{\textcolor{numpertdeltaimprove}{+4.70}} & 88.43\textsuperscript{\textcolor{numpertdeltaimprove}{+3.20}} & 88.42\textsuperscript{\textcolor{numpertdeltaimprove}{+3.19}} & 98.08\textsuperscript{\textcolor{numpertdeltaimprove}{+12.85}} & 99.54\textsuperscript{\textcolor{numpertdeltaimprove}{+14.31}} & 94.91\textsuperscript{\textcolor{numpertdeltaimprove}{+9.68}} & 92.81\textsuperscript{\textcolor{numpertdeltaimprove}{+7.58}} \\
 & \mallperturbfour & 86.36 & \textbf{100.00}\textsuperscript{\textcolor{numpertdeltaimprove}{+13.64}} & 95.37\textsuperscript{\textcolor{numpertdeltaimprove}{+9.01}} & 98.42\textsuperscript{\textcolor{numpertdeltaimprove}{+12.06}} & \textbf{100.00}\textsuperscript{\textcolor{numpertdeltaimprove}{+13.64}} & \textbf{100.00}\textsuperscript{\textcolor{numpertdeltaimprove}{+13.64}} & 98.61\textsuperscript{\textcolor{numpertdeltaimprove}{+12.25}} & 98.44\textsuperscript{\textcolor{numpertdeltaimprove}{+12.08}} \\
 & \mallperturb & 92.42 & 97.84\textsuperscript{\textcolor{numpertdeltaimprove}{+5.42}} & 96.76\textsuperscript{\textcolor{numpertdeltaimprove}{+4.34}} & 96.84\textsuperscript{\textcolor{numpertdeltaimprove}{+4.42}} & \textbf{100.00}\textsuperscript{\textcolor{numpertdeltaimprove}{+7.58}} & \textbf{100.00}\textsuperscript{\textcolor{numpertdeltaimprove}{+7.58}} & 99.54\textsuperscript{\textcolor{numpertdeltaimprove}{+7.12}} & 98.35\textsuperscript{\textcolor{numpertdeltaimprove}{+5.93}} \\
\midrule
\Block{6-1}{\rotate \groupthreetable} & \mapproximation & 92.05 & 96.40\textsuperscript{\textcolor{numpertdeltaimprove}{+4.35}} & 36.11\textsuperscript{\textcolor{numpertdeltadecline}{-55.94}} & 85.79\textsuperscript{\textcolor{numpertdeltadecline}{-6.26}} & 21.15\textsuperscript{\textcolor{numpertdeltadecline}{-70.90}} & 36.57\textsuperscript{\textcolor{numpertdeltadecline}{-55.48}} & 92.13\textsuperscript{\textcolor{numpertdeltaimprove}{+0.08}} & 64.53\textsuperscript{\textcolor{numpertdeltadecline}{-27.52}} \\
 & \mnumeration & 92.42 & 92.81\textsuperscript{\textcolor{numpertdeltaimprove}{+0.39}} & 98.15\textsuperscript{\textcolor{numpertdeltaimprove}{+5.73}} & 96.84\textsuperscript{\textcolor{numpertdeltaimprove}{+4.42}} & 13.46\textsuperscript{\textcolor{numpertdeltadecline}{-78.96}} & 28.24\textsuperscript{\textcolor{numpertdeltadecline}{-64.18}} & 95.83\textsuperscript{\textcolor{numpertdeltaimprove}{+3.41}} & 77.75\textsuperscript{\textcolor{numpertdeltadecline}{-14.67}} \\
 & \mrange & 92.05 & 72.66\textsuperscript{\textcolor{numpertdeltadecline}{-19.39}} & 42.13\textsuperscript{\textcolor{numpertdeltadecline}{-49.92}} & 97.37\textsuperscript{\textcolor{numpertdeltaimprove}{+5.32}} & 19.23\textsuperscript{\textcolor{numpertdeltadecline}{-72.82}} & 11.11\textsuperscript{\textcolor{numpertdeltadecline}{-80.94}} & 78.70\textsuperscript{\textcolor{numpertdeltadecline}{-13.35}} & 56.46\textsuperscript{\textcolor{numpertdeltadecline}{-35.59}} \\
 & \mnegation & 88.64 & 54.68\textsuperscript{\textcolor{numpertdeltadecline}{-33.96}} & 36.57\textsuperscript{\textcolor{numpertdeltadecline}{-52.07}} & 50.53\textsuperscript{\textcolor{numpertdeltadecline}{-38.11}} & 98.08\textsuperscript{\textcolor{numpertdeltaimprove}{+9.44}} & 19.91\textsuperscript{\textcolor{numpertdeltadecline}{-68.73}} & 68.98\textsuperscript{\textcolor{numpertdeltadecline}{-19.66}} & 48.01\textsuperscript{\textcolor{numpertdeltadecline}{-40.63}} \\
 & \mmasked & 90.20 & 29.01\textsuperscript{\textcolor{numpertdeltadecline}{-61.19}} & 23.38\textsuperscript{\textcolor{numpertdeltadecline}{-66.82}} & 33.15\textsuperscript{\textcolor{numpertdeltadecline}{-57.05}} & 27.08\textsuperscript{\textcolor{numpertdeltadecline}{-63.12}} & 99.50\textsuperscript{\textcolor{numpertdeltaimprove}{+9.30}} & 42.05\textsuperscript{\textcolor{numpertdeltadecline}{-48.15}} & 46.03\textsuperscript{\textcolor{numpertdeltadecline}{-44.17}} \\
 & \mrandom & 92.24 & 97.62\textsuperscript{\textcolor{numpertdeltaimprove}{+5.38}} & 74.87\textsuperscript{\textcolor{numpertdeltadecline}{-17.37}} & 94.61\textsuperscript{\textcolor{numpertdeltaimprove}{+2.37}} & 38.78\textsuperscript{\textcolor{numpertdeltadecline}{-53.46}} & 59.30\textsuperscript{\textcolor{numpertdeltadecline}{-32.94}} & 98.41\textsuperscript{\textcolor{numpertdeltaimprove}{+6.17}} & 81.44\textsuperscript{\textcolor{numpertdeltadecline}{-10.80}} \\
\midrule
\Block{7-1}{\rotate \groupfourtable} & \mnooriginal & 93.56 & 98.56\textsuperscript{\textcolor{numpertdeltaimprove}{+5.00}} & \textbf{99.07}\textsuperscript{\textcolor{numpertdeltaimprove}{+5.51}} & 98.42\textsuperscript{\textcolor{numpertdeltaimprove}{+4.86}} & \textbf{100.00}\textsuperscript{\textcolor{numpertdeltaimprove}{+6.44}} & \textbf{100.00}\textsuperscript{\textcolor{numpertdeltaimprove}{+6.44}} & 98.61\textsuperscript{\textcolor{numpertdeltaimprove}{+5.05}} & \textbf{99.03}\textsuperscript{\textcolor{numpertdeltaimprove}{+5.47}} \\
 & \mnoapproximation & 88.64 & 99.28\textsuperscript{\textcolor{numpertdeltaimprove}{+10.64}} & 98.15\textsuperscript{\textcolor{numpertdeltaimprove}{+9.51}} & 97.37\textsuperscript{\textcolor{numpertdeltaimprove}{+8.73}} & \textbf{100.00}\textsuperscript{\textcolor{numpertdeltaimprove}{+11.36}} & \textbf{100.00}\textsuperscript{\textcolor{numpertdeltaimprove}{+11.36}} & 99.54\textsuperscript{\textcolor{numpertdeltaimprove}{+10.90}} & 98.93\textsuperscript{\textcolor{numpertdeltaimprove}{+10.29}} \\
 & \mnonumeration & 89.39 & 97.12\textsuperscript{\textcolor{numpertdeltaimprove}{+7.73}} & 67.59\textsuperscript{\textcolor{numpertdeltadecline}{-21.80}} & 96.84\textsuperscript{\textcolor{numpertdeltaimprove}{+7.45}} & \textbf{100.00}\textsuperscript{\textcolor{numpertdeltaimprove}{+10.61}} & \textbf{100.00}\textsuperscript{\textcolor{numpertdeltaimprove}{+10.61}} & 98.61\textsuperscript{\textcolor{numpertdeltaimprove}{+9.22}} & 91.93\textsuperscript{\textcolor{numpertdeltaimprove}{+2.54}} \\
 & \mnorange & 92.80 & 94.96\textsuperscript{\textcolor{numpertdeltaimprove}{+2.16}} & 95.83\textsuperscript{\textcolor{numpertdeltaimprove}{+3.03}} & 95.79\textsuperscript{\textcolor{numpertdeltaimprove}{+2.99}} & 98.08\textsuperscript{\textcolor{numpertdeltaimprove}{+5.28}} & 99.54\textsuperscript{\textcolor{numpertdeltaimprove}{+6.74}} & 97.69\textsuperscript{\textcolor{numpertdeltaimprove}{+4.89}} & 96.99\textsuperscript{\textcolor{numpertdeltaimprove}{+4.19}} \\
 & \mnonegation & 89.02 & 99.28\textsuperscript{\textcolor{numpertdeltaimprove}{+10.26}} & 98.61\textsuperscript{\textcolor{numpertdeltaimprove}{+9.59}} & 97.37\textsuperscript{\textcolor{numpertdeltaimprove}{+8.35}} & 65.38\textsuperscript{\textcolor{numpertdeltadecline}{-23.64}} & \textbf{100.00}\textsuperscript{\textcolor{numpertdeltaimprove}{+10.98}} & 97.69\textsuperscript{\textcolor{numpertdeltaimprove}{+8.67}} & 96.89\textsuperscript{\textcolor{numpertdeltaimprove}{+7.87}} \\
 & \mnomasked & 92.05 & 95.68\textsuperscript{\textcolor{numpertdeltaimprove}{+3.63}} & 96.76\textsuperscript{\textcolor{numpertdeltaimprove}{+4.71}} & 97.89\textsuperscript{\textcolor{numpertdeltaimprove}{+5.84}} & 98.08\textsuperscript{\textcolor{numpertdeltaimprove}{+6.03}} & 16.20\textsuperscript{\textcolor{numpertdeltadecline}{-75.85}} & 98.15\textsuperscript{\textcolor{numpertdeltaimprove}{+6.10}} & 80.27\textsuperscript{\textcolor{numpertdeltadecline}{-11.78}} \\
 & \mnorandom & 94.32 & 94.24\textsuperscript{\textcolor{numpertdeltadecline}{-0.08}} & 95.83\textsuperscript{\textcolor{numpertdeltaimprove}{+1.51}} & 95.26\textsuperscript{\textcolor{numpertdeltaimprove}{+0.94}} & 98.08\textsuperscript{\textcolor{numpertdeltaimprove}{+3.76}} & 99.54\textsuperscript{\textcolor{numpertdeltaimprove}{+5.22}} & 95.83\textsuperscript{\textcolor{numpertdeltaimprove}{+1.51}} & 96.40\textsuperscript{\textcolor{numpertdeltaimprove}{+2.08}} \\
\bottomrule
\end{NiceTabular}
}
\caption{\footnotesize \ttf\ accuracy (\%) for Qwen3-8B (unquantized fp16), selected models. Columns as in Table~\ref{tab:ttf}.}
\label{tab:unquant-ttf}
\end{table*}

%% file: appendix_tables_figures/appendix_results_unquant/T-_T.tex
\begin{table*}[t!!!]
\centering
\definecolor{numpertdeltaimprove}{RGB}{0,100,0}
\definecolor{numpertdeltadecline}{RGB}{180,35,24}
\small
\begin{NiceTabular}{l|ccccc}[first-col]
\toprule
 & \textbf{Model} & \textbf{Orig.} & \textbf{Approx.} & \textbf{Num.} & \textbf{Range} & \textbf{Micro Avg.} \\
\midrule
\Block{5-1}{\rotate \grouponetable} & \mbase & 94.35 & 93.67\textsuperscript{\textcolor{numpertdeltadecline}{-0.68}} & 97.33\textsuperscript{\textcolor{numpertdeltaimprove}{+2.98}} & \textbf{98.17}\textsuperscript{\textcolor{numpertdeltaimprove}{+3.82}} & \textbf{96.69}\textsuperscript{\textcolor{numpertdeltaimprove}{+2.34}} \\
 & \mbasetwoshot & 86.92 & 80.83\textsuperscript{\textcolor{numpertdeltadecline}{-6.09}} & 86.14\textsuperscript{\textcolor{numpertdeltadecline}{-0.78}} & 89.41\textsuperscript{\textcolor{numpertdeltaimprove}{+2.49}} & 85.93\textsuperscript{\textcolor{numpertdeltadecline}{-0.99}} \\
 & \mbasethirtytwo & 88.24 & 89.29\textsuperscript{\textcolor{numpertdeltaimprove}{+1.05}} & 90.37\textsuperscript{\textcolor{numpertdeltaimprove}{+2.13}} & 87.84\textsuperscript{\textcolor{numpertdeltadecline}{-0.40}} & 89.21\textsuperscript{\textcolor{numpertdeltaimprove}{+0.97}} \\
 & \mnoperturb & \textbf{94.70} & 95.68\textsuperscript{\textcolor{numpertdeltaimprove}{+0.98}} & 95.37\textsuperscript{\textcolor{numpertdeltaimprove}{+0.67}} & 94.21\textsuperscript{\textcolor{numpertdeltadecline}{-0.49}} & 95.04\textsuperscript{\textcolor{numpertdeltaimprove}{+0.34}} \\
\midrule
\Block{4-1}{\rotate \grouptwotable} & \mallperturbsix & 81.06 & 84.17\textsuperscript{\textcolor{numpertdeltaimprove}{+3.11}} & 78.70\textsuperscript{\textcolor{numpertdeltadecline}{-2.36}} & 87.37\textsuperscript{\textcolor{numpertdeltaimprove}{+6.31}} & 83.12\textsuperscript{\textcolor{numpertdeltaimprove}{+2.06}} \\
 & \mallperturbseventeen & 85.23 & 89.93\textsuperscript{\textcolor{numpertdeltaimprove}{+4.70}} & 85.65\textsuperscript{\textcolor{numpertdeltaimprove}{+0.42}} & 91.58\textsuperscript{\textcolor{numpertdeltaimprove}{+6.35}} & 88.81\textsuperscript{\textcolor{numpertdeltaimprove}{+3.58}} \\
 & \mallperturbfour & 86.36 & 89.21\textsuperscript{\textcolor{numpertdeltaimprove}{+2.85}} & 88.89\textsuperscript{\textcolor{numpertdeltaimprove}{+2.53}} & 90.00\textsuperscript{\textcolor{numpertdeltaimprove}{+3.64}} & 89.36\textsuperscript{\textcolor{numpertdeltaimprove}{+3.00}} \\
 & \mallperturb & 92.42 & 87.77\textsuperscript{\textcolor{numpertdeltadecline}{-4.65}} & 90.74\textsuperscript{\textcolor{numpertdeltadecline}{-1.68}} & 92.63\textsuperscript{\textcolor{numpertdeltaimprove}{+0.21}} & 90.64\textsuperscript{\textcolor{numpertdeltadecline}{-1.78}} \\
\midrule
\Block{6-1}{\rotate \groupthreetable} & \mapproximation & 92.05 & 91.37\textsuperscript{\textcolor{numpertdeltadecline}{-0.68}} & 93.06\textsuperscript{\textcolor{numpertdeltaimprove}{+1.01}} & 74.74\textsuperscript{\textcolor{numpertdeltadecline}{-17.31}} & 86.24\textsuperscript{\textcolor{numpertdeltadecline}{-5.81}} \\
 & \mnumeration & 92.42 & 71.94\textsuperscript{\textcolor{numpertdeltadecline}{-20.48}} & 93.52\textsuperscript{\textcolor{numpertdeltaimprove}{+1.10}} & 27.37\textsuperscript{\textcolor{numpertdeltadecline}{-65.05}} & 64.95\textsuperscript{\textcolor{numpertdeltadecline}{-27.47}} \\
 & \mrange & 92.05 & 92.81\textsuperscript{\textcolor{numpertdeltaimprove}{+0.76}} & 92.13\textsuperscript{\textcolor{numpertdeltaimprove}{+0.08}} & 93.68\textsuperscript{\textcolor{numpertdeltaimprove}{+1.63}} & 92.84\textsuperscript{\textcolor{numpertdeltaimprove}{+0.79}} \\
 & \mnegation & 88.64 & 84.89\textsuperscript{\textcolor{numpertdeltadecline}{-3.75}} & 90.74\textsuperscript{\textcolor{numpertdeltaimprove}{+2.10}} & 76.32\textsuperscript{\textcolor{numpertdeltadecline}{-12.32}} & 84.22\textsuperscript{\textcolor{numpertdeltadecline}{-4.42}} \\
 & \mmasked & 90.20 & 89.63\textsuperscript{\textcolor{numpertdeltadecline}{-0.57}} & 89.66\textsuperscript{\textcolor{numpertdeltadecline}{-0.54}} & 87.43\textsuperscript{\textcolor{numpertdeltadecline}{-2.77}} & 88.87\textsuperscript{\textcolor{numpertdeltadecline}{-1.33}} \\
 & \mrandom & 92.24 & 58.27\textsuperscript{\textcolor{numpertdeltadecline}{-33.97}} & 91.54\textsuperscript{\textcolor{numpertdeltadecline}{-0.70}} & 23.56\textsuperscript{\textcolor{numpertdeltadecline}{-68.68}} & 59.36\textsuperscript{\textcolor{numpertdeltadecline}{-32.88}} \\
\midrule
\Block{7-1}{\rotate \groupfourtable} & \mnooriginal & 93.56 & 94.24\textsuperscript{\textcolor{numpertdeltaimprove}{+0.68}} & 93.06\textsuperscript{\textcolor{numpertdeltadecline}{-0.50}} & 94.74\textsuperscript{\textcolor{numpertdeltaimprove}{+1.18}} & 93.95\textsuperscript{\textcolor{numpertdeltaimprove}{+0.39}} \\
 & \mnoapproximation & 88.64 & 55.40\textsuperscript{\textcolor{numpertdeltadecline}{-33.24}} & 87.04\textsuperscript{\textcolor{numpertdeltadecline}{-1.60}} & 90.53\textsuperscript{\textcolor{numpertdeltaimprove}{+1.89}} & 80.19\textsuperscript{\textcolor{numpertdeltadecline}{-8.45}} \\
 & \mnonumeration & 89.39 & 92.81\textsuperscript{\textcolor{numpertdeltaimprove}{+3.42}} & 87.50\textsuperscript{\textcolor{numpertdeltadecline}{-1.89}} & 93.16\textsuperscript{\textcolor{numpertdeltaimprove}{+3.77}} & 90.83\textsuperscript{\textcolor{numpertdeltaimprove}{+1.44}} \\
 & \mnorange & 92.80 & 94.96\textsuperscript{\textcolor{numpertdeltaimprove}{+2.16}} & 93.98\textsuperscript{\textcolor{numpertdeltaimprove}{+1.18}} & 52.11\textsuperscript{\textcolor{numpertdeltadecline}{-40.69}} & 79.63\textsuperscript{\textcolor{numpertdeltadecline}{-13.17}} \\
 & \mnonegation & 89.02 & 92.09\textsuperscript{\textcolor{numpertdeltaimprove}{+3.07}} & 86.57\textsuperscript{\textcolor{numpertdeltadecline}{-2.45}} & 90.53\textsuperscript{\textcolor{numpertdeltaimprove}{+1.51}} & 89.36\textsuperscript{\textcolor{numpertdeltaimprove}{+0.34}} \\
 & \mnomasked & 92.05 & 93.53\textsuperscript{\textcolor{numpertdeltaimprove}{+1.48}} & 92.59\textsuperscript{\textcolor{numpertdeltaimprove}{+0.54}} & 92.63\textsuperscript{\textcolor{numpertdeltaimprove}{+0.58}} & 92.84\textsuperscript{\textcolor{numpertdeltaimprove}{+0.79}} \\
 & \mnorandom & 94.32 & 96.40\textsuperscript{\textcolor{numpertdeltaimprove}{+2.08}} & 92.13\textsuperscript{\textcolor{numpertdeltadecline}{-2.19}} & 94.74\textsuperscript{\textcolor{numpertdeltaimprove}{+0.42}} & 94.13\textsuperscript{\textcolor{numpertdeltadecline}{-0.19}} \\
\bottomrule
\end{NiceTabular}
\caption{\footnotesize \ttt\ accuracy (\%) for Qwen3-8B (unquantised, BF16/FP16), selected models. Columns as in Table~\ref{tab:ttt}.}
\label{tab:unquant-ttt}
\end{table*}

%% file: appendix_tables_figures/appendix_results_unquant/F-_F.tex
\begin{table*}[t!!!]
\centering
\definecolor{numpertdeltaimprove}{RGB}{0,100,0}
\definecolor{numpertdeltadecline}{RGB}{180,35,24}
\small
\setlength{\tabcolsep}{3pt}
\resizebox{\textwidth}{!}{%
\begin{NiceTabular}{l|cccccccc}[first-col]
\toprule
 & \textbf{Model} & \textbf{Orig.} & \textbf{Approx.} & \textbf{Num.} & \textbf{Range} & \textbf{Neg.} & \textbf{Mask.} & \textbf{Rand.} & \textbf{Micro Avg.} \\
\midrule
\Block{5-1}{\rotate \grouponetable} & \mbase & 91.69 & 95.73\textsuperscript{\textcolor{numpertdeltaimprove}{+4.04}} & 94.37\textsuperscript{\textcolor{numpertdeltaimprove}{+2.68}} & 86.88\textsuperscript{\textcolor{numpertdeltadecline}{-4.81}} & 88.10\textsuperscript{\textcolor{numpertdeltadecline}{-3.59}} & 92.52\textsuperscript{\textcolor{numpertdeltaimprove}{+0.83}} & 95.91\textsuperscript{\textcolor{numpertdeltaimprove}{+4.22}} & 92.98\textsuperscript{\textcolor{numpertdeltaimprove}{+1.29}} \\
 & \mbasetwoshot & 94.95 & 96.77\textsuperscript{\textcolor{numpertdeltaimprove}{+1.82}} & 96.30\textsuperscript{\textcolor{numpertdeltaimprove}{+1.35}} & 92.31\textsuperscript{\textcolor{numpertdeltadecline}{-2.64}} & 93.10\textsuperscript{\textcolor{numpertdeltadecline}{-1.85}} & 93.62\textsuperscript{\textcolor{numpertdeltadecline}{-1.33}} & 98.33\textsuperscript{\textcolor{numpertdeltaimprove}{+3.38}} & 95.42\textsuperscript{\textcolor{numpertdeltaimprove}{+0.47}} \\
 & \mbasethirtytwo & 95.99 & 96.69\textsuperscript{\textcolor{numpertdeltaimprove}{+0.70}} & 95.87\textsuperscript{\textcolor{numpertdeltadecline}{-0.12}} & 94.48\textsuperscript{\textcolor{numpertdeltadecline}{-1.51}} & 93.75\textsuperscript{\textcolor{numpertdeltadecline}{-2.24}} & 95.50\textsuperscript{\textcolor{numpertdeltadecline}{-0.49}} & 98.21\textsuperscript{\textcolor{numpertdeltaimprove}{+2.22}} & 96.09\textsuperscript{\textcolor{numpertdeltaimprove}{+0.10}} \\
 & \mnoperturb & 97.56 & 97.45\textsuperscript{\textcolor{numpertdeltadecline}{-0.11}} & 97.21\textsuperscript{\textcolor{numpertdeltadecline}{-0.35}} & 96.92\textsuperscript{\textcolor{numpertdeltadecline}{-0.64}} & 95.60\textsuperscript{\textcolor{numpertdeltadecline}{-1.96}} & 96.81\textsuperscript{\textcolor{numpertdeltadecline}{-0.75}} & 98.40\textsuperscript{\textcolor{numpertdeltaimprove}{+0.84}} & 97.30\textsuperscript{\textcolor{numpertdeltadecline}{-0.26}} \\
\midrule
\Block{4-1}{\rotate \grouptwotable} & \mallperturbsix & 95.93 & 95.15\textsuperscript{\textcolor{numpertdeltadecline}{-0.78}} & 95.41\textsuperscript{\textcolor{numpertdeltadecline}{-0.52}} & 88.86\textsuperscript{\textcolor{numpertdeltadecline}{-7.07}} & \textbf{100.00}\textsuperscript{\textcolor{numpertdeltaimprove}{+4.07}} & \textbf{100.00}\textsuperscript{\textcolor{numpertdeltaimprove}{+4.07}} & \textbf{100.00}\textsuperscript{\textcolor{numpertdeltaimprove}{+4.07}} & 96.30\textsuperscript{\textcolor{numpertdeltaimprove}{+0.37}} \\
 & \mallperturbseventeen & 96.59 & 94.64\textsuperscript{\textcolor{numpertdeltadecline}{-1.95}} & 95.21\textsuperscript{\textcolor{numpertdeltadecline}{-1.38}} & 92.18\textsuperscript{\textcolor{numpertdeltadecline}{-4.41}} & \textbf{100.00}\textsuperscript{\textcolor{numpertdeltaimprove}{+3.41}} & \textbf{100.00}\textsuperscript{\textcolor{numpertdeltaimprove}{+3.41}} & 99.40\textsuperscript{\textcolor{numpertdeltaimprove}{+2.81}} & 96.64\textsuperscript{\textcolor{numpertdeltaimprove}{+0.05}} \\
 & \mallperturbfour & \textbf{98.37} & 98.21\textsuperscript{\textcolor{numpertdeltadecline}{-0.16}} & 97.60\textsuperscript{\textcolor{numpertdeltadecline}{-0.77}} & 94.79\textsuperscript{\textcolor{numpertdeltadecline}{-3.58}} & \textbf{100.00}\textsuperscript{\textcolor{numpertdeltaimprove}{+1.63}} & \textbf{100.00}\textsuperscript{\textcolor{numpertdeltaimprove}{+1.63}} & \textbf{100.00}\textsuperscript{\textcolor{numpertdeltaimprove}{+1.63}} & 98.30\textsuperscript{\textcolor{numpertdeltadecline}{-0.07}} \\
 & \mallperturb & 98.21 & 96.43\textsuperscript{\textcolor{numpertdeltadecline}{-1.78}} & 97.41\textsuperscript{\textcolor{numpertdeltadecline}{-0.80}} & 94.08\textsuperscript{\textcolor{numpertdeltadecline}{-4.13}} & \textbf{100.00}\textsuperscript{\textcolor{numpertdeltaimprove}{+1.79}} & \textbf{100.00}\textsuperscript{\textcolor{numpertdeltaimprove}{+1.79}} & \textbf{100.00}\textsuperscript{\textcolor{numpertdeltaimprove}{+1.79}} & 97.84\textsuperscript{\textcolor{numpertdeltadecline}{-0.37}} \\
\midrule
\Block{6-1}{\rotate \groupthreetable} & \mapproximation & 97.24 & 95.66\textsuperscript{\textcolor{numpertdeltadecline}{-1.58}} & 97.01\textsuperscript{\textcolor{numpertdeltadecline}{-0.23}} & 96.68\textsuperscript{\textcolor{numpertdeltadecline}{-0.56}} & 95.60\textsuperscript{\textcolor{numpertdeltadecline}{-1.64}} & 98.00\textsuperscript{\textcolor{numpertdeltaimprove}{+0.76}} & 99.40\textsuperscript{\textcolor{numpertdeltaimprove}{+2.16}} & 97.38\textsuperscript{\textcolor{numpertdeltaimprove}{+0.14}} \\
 & \mnumeration & 97.40 & 98.47\textsuperscript{\textcolor{numpertdeltaimprove}{+1.07}} & 97.21\textsuperscript{\textcolor{numpertdeltadecline}{-0.19}} & 99.53\textsuperscript{\textcolor{numpertdeltaimprove}{+2.13}} & 95.60\textsuperscript{\textcolor{numpertdeltadecline}{-1.80}} & 97.41\textsuperscript{\textcolor{numpertdeltaimprove}{+0.01}} & \textbf{100.00}\textsuperscript{\textcolor{numpertdeltaimprove}{+2.60}} & 98.38\textsuperscript{\textcolor{numpertdeltaimprove}{+0.98}} \\
 & \mrange & 97.56 & 97.19\textsuperscript{\textcolor{numpertdeltadecline}{-0.37}} & 97.60\textsuperscript{\textcolor{numpertdeltaimprove}{+0.04}} & 94.31\textsuperscript{\textcolor{numpertdeltadecline}{-3.25}} & 97.80\textsuperscript{\textcolor{numpertdeltaimprove}{+0.24}} & 97.21\textsuperscript{\textcolor{numpertdeltadecline}{-0.35}} & 99.60\textsuperscript{\textcolor{numpertdeltaimprove}{+2.04}} & 97.30\textsuperscript{\textcolor{numpertdeltadecline}{-0.26}} \\
 & \mnegation & 96.42 & 97.70\textsuperscript{\textcolor{numpertdeltaimprove}{+1.28}} & 97.41\textsuperscript{\textcolor{numpertdeltaimprove}{+0.99}} & 98.10\textsuperscript{\textcolor{numpertdeltaimprove}{+1.68}} & \textbf{100.00}\textsuperscript{\textcolor{numpertdeltaimprove}{+3.58}} & 96.81\textsuperscript{\textcolor{numpertdeltaimprove}{+0.39}} & 99.00\textsuperscript{\textcolor{numpertdeltaimprove}{+2.58}} & 97.88\textsuperscript{\textcolor{numpertdeltaimprove}{+1.46}} \\
 & \mmasked & 97.31 & 97.85\textsuperscript{\textcolor{numpertdeltaimprove}{+0.54}} & 97.50\textsuperscript{\textcolor{numpertdeltaimprove}{+0.19}} & 97.27\textsuperscript{\textcolor{numpertdeltadecline}{-0.04}} & 95.40\textsuperscript{\textcolor{numpertdeltadecline}{-1.91}} & \textbf{100.00}\textsuperscript{\textcolor{numpertdeltaimprove}{+2.69}} & 99.16\textsuperscript{\textcolor{numpertdeltaimprove}{+1.85}} & 98.30\textsuperscript{\textcolor{numpertdeltaimprove}{+0.99}} \\
 & \mrandom & 96.22 & 98.57\textsuperscript{\textcolor{numpertdeltaimprove}{+2.35}} & 97.22\textsuperscript{\textcolor{numpertdeltaimprove}{+1.00}} & \textbf{100.00}\textsuperscript{\textcolor{numpertdeltaimprove}{+3.78}} & 98.73\textsuperscript{\textcolor{numpertdeltaimprove}{+2.51}} & 99.10\textsuperscript{\textcolor{numpertdeltaimprove}{+2.88}} & \textbf{100.00}\textsuperscript{\textcolor{numpertdeltaimprove}{+3.78}} & \textbf{98.95}\textsuperscript{\textcolor{numpertdeltaimprove}{+2.73}} \\
\midrule
\Block{7-1}{\rotate \groupfourtable} & \mnooriginal & 96.91 & 95.66\textsuperscript{\textcolor{numpertdeltadecline}{-1.25}} & 97.01\textsuperscript{\textcolor{numpertdeltaimprove}{+0.10}} & 94.79\textsuperscript{\textcolor{numpertdeltadecline}{-2.12}} & \textbf{100.00}\textsuperscript{\textcolor{numpertdeltaimprove}{+3.09}} & \textbf{100.00}\textsuperscript{\textcolor{numpertdeltaimprove}{+3.09}} & \textbf{100.00}\textsuperscript{\textcolor{numpertdeltaimprove}{+3.09}} & 97.76\textsuperscript{\textcolor{numpertdeltaimprove}{+0.85}} \\
 & \mnoapproximation & 98.05 & 98.98\textsuperscript{\textcolor{numpertdeltaimprove}{+0.93}} & 97.80\textsuperscript{\textcolor{numpertdeltadecline}{-0.25}} & 95.26\textsuperscript{\textcolor{numpertdeltadecline}{-2.79}} & \textbf{100.00}\textsuperscript{\textcolor{numpertdeltaimprove}{+1.95}} & \textbf{100.00}\textsuperscript{\textcolor{numpertdeltaimprove}{+1.95}} & \textbf{100.00}\textsuperscript{\textcolor{numpertdeltaimprove}{+1.95}} & 98.55\textsuperscript{\textcolor{numpertdeltaimprove}{+0.50}} \\
 & \mnonumeration & 98.21 & 97.45\textsuperscript{\textcolor{numpertdeltadecline}{-0.76}} & 97.21\textsuperscript{\textcolor{numpertdeltadecline}{-1.00}} & 93.84\textsuperscript{\textcolor{numpertdeltadecline}{-4.37}} & \textbf{100.00}\textsuperscript{\textcolor{numpertdeltaimprove}{+1.79}} & \textbf{100.00}\textsuperscript{\textcolor{numpertdeltaimprove}{+1.79}} & \textbf{100.00}\textsuperscript{\textcolor{numpertdeltaimprove}{+1.79}} & 97.92\textsuperscript{\textcolor{numpertdeltadecline}{-0.29}} \\
 & \mnorange & 97.07 & 96.43\textsuperscript{\textcolor{numpertdeltadecline}{-0.64}} & 96.41\textsuperscript{\textcolor{numpertdeltadecline}{-0.66}} & 98.58\textsuperscript{\textcolor{numpertdeltaimprove}{+1.51}} & \textbf{100.00}\textsuperscript{\textcolor{numpertdeltaimprove}{+2.93}} & \textbf{100.00}\textsuperscript{\textcolor{numpertdeltaimprove}{+2.93}} & \textbf{100.00}\textsuperscript{\textcolor{numpertdeltaimprove}{+2.93}} & 98.42\textsuperscript{\textcolor{numpertdeltaimprove}{+1.35}} \\
 & \mnonegation & \textbf{98.37} & 97.96\textsuperscript{\textcolor{numpertdeltadecline}{-0.41}} & 98.40\textsuperscript{\textcolor{numpertdeltaimprove}{+0.03}} & 97.63\textsuperscript{\textcolor{numpertdeltadecline}{-0.74}} & 97.80\textsuperscript{\textcolor{numpertdeltadecline}{-0.57}} & \textbf{100.00}\textsuperscript{\textcolor{numpertdeltaimprove}{+1.63}} & \textbf{100.00}\textsuperscript{\textcolor{numpertdeltaimprove}{+1.63}} & 98.84\textsuperscript{\textcolor{numpertdeltaimprove}{+0.47}} \\
 & \mnomasked & 97.24 & 96.43\textsuperscript{\textcolor{numpertdeltadecline}{-0.81}} & 97.60\textsuperscript{\textcolor{numpertdeltaimprove}{+0.36}} & 94.08\textsuperscript{\textcolor{numpertdeltadecline}{-3.16}} & \textbf{100.00}\textsuperscript{\textcolor{numpertdeltaimprove}{+2.76}} & 97.01\textsuperscript{\textcolor{numpertdeltadecline}{-0.23}} & \textbf{100.00}\textsuperscript{\textcolor{numpertdeltaimprove}{+2.76}} & 97.26\textsuperscript{\textcolor{numpertdeltaimprove}{+0.02}} \\
 & \mnorandom & 96.91 & 95.92\textsuperscript{\textcolor{numpertdeltadecline}{-0.99}} & 96.21\textsuperscript{\textcolor{numpertdeltadecline}{-0.70}} & 94.08\textsuperscript{\textcolor{numpertdeltadecline}{-2.83}} & \textbf{100.00}\textsuperscript{\textcolor{numpertdeltaimprove}{+3.09}} & \textbf{100.00}\textsuperscript{\textcolor{numpertdeltaimprove}{+3.09}} & \textbf{100.00}\textsuperscript{\textcolor{numpertdeltaimprove}{+3.09}} & 97.51\textsuperscript{\textcolor{numpertdeltaimprove}{+0.60}} \\
\bottomrule
\end{NiceTabular}
}
\caption{\footnotesize \ftf\ accuracy (\%) for Qwen3-8B (unquantized fp16), selected models. Columns as in Table~\ref{tab:ftf}}
\label{tab:unquant-ftf}
\end{table*}